\documentclass[11pt,a4paper]{article}

\usepackage[utf8]{inputenc}
\usepackage[T1]{fontenc}
\usepackage[british]{babel}
\usepackage[margin=2.6cm]{geometry}
\usepackage{graphicx}
\usepackage{amsmath,amssymb}
\usepackage{booktabs}
\usepackage{url}
\usepackage{xcolor}
\usepackage{enumitem}
\usepackage{microtype}
\usepackage{cite}
\usepackage{tikz}
\usetikzlibrary{arrows.meta, positioning, shapes.geometric, fit, backgrounds}
\usepackage[font=small,labelfont=bf]{caption}
\setlist[itemize]{leftmargin=1.5em,itemsep=2pt,topsep=3pt}

\title{\bfseries When Agents Coordinate:\\ Measuring Coordination in Multi-Agent AI Coding}
\author{Giuseppe Destefanis\thanks{Corresponding author: \texttt{g.destefanis@ucl.ac.uk}}\ \ and Tomaso Aste\\
\normalsize Department of Computer Science, University College London, UK}
\date{}

\begin{document}
\maketitle

\begin{abstract}
\noindent
We study how teams of AI coding agents coordinate while solving programming tasks. Current evaluations usually report whether the agents complete the task and how much the run costs, leaving the coordination inside the team largely unmeasured. We introduce an instrument to measure this coordination directly. Each run is represented as a temporal network in which agents and files are nodes, and messages, file writes, and file reads are timestamped directed edges with an associated cost. We apply this instrument to 1{,}902 runs, each evaluated with a fixed test suite, across configurations that vary the team size, the team structure, and the file policy.

The resulting networks show how coordination changes as teams grow and as the work changes. Direct messaging initially increases close to quadratically with the number of agents, with much of this growth coming from an early round of introductions. As the teams grow further, this increase levels off in the largest teams we study, where agents increasingly communicate through broadcast messages. The programming task also shapes the network that emerges. Work built around a shared specification produces dense, highly connected teams, while pipeline tasks produce sparse networks organised around local interfaces. Shared files can replace repeated one-to-one communication, cutting output tokens by about 42\% at eight agents on message-heavy work, while adding overhead when files already carry the coordination. Naming one agent as coordinator creates no communication hub and provides no reliable improvement in success.

We also observe an unprompted tendency for agents to seek out hidden grading material. We repeat the key experimental conditions in a sealed environment, replacing the hidden material with marked placeholder files. Across 244 additional runs, agents still reach for it in four fifths of runs, while the coordinator and file-channel findings reproduce.

Repeated runs also show that coordination measurements can vary substantially under the same configuration and pinned model, especially when the task leaves agents more freedom in how to organise. A single run therefore gives only one sample of the coordination behaviour of a configuration. These results show that coordination among AI coding agents can be measured directly, exposing structures, behaviours, and costs that task success and aggregate token counts leave hidden.
\end{abstract}

\vspace{4pt}
\noindent\textbf{Keywords:} multi-agent systems, AI agents, large language models, software engineering, network analysis, empirical study

\section{Introduction}
\label{sec:introduction}

Multi-agent AI coding systems assign a programming task to several agents that work in parallel in a shared workspace. Whether the team succeeds, and what the run costs, depends on how the agents coordinate: how they divide the work, what they tell each other, which files they share, and how they resolve the interfaces between their contributions. Almost none of this survives into the final code. A run can pass every test while leaving no trace of how much coordination it required, and two runs that pass the same tests can differ severalfold in messages, file activity, and tokens.

A developer configuring such a system must decide the team size, whether to name one agent coordinator, and whether the team should coordinate through direct messaging or through shared files. The effects of these choices are still poorly understood. Evaluations typically report task success and the final code or patch, with computational cost increasingly considered as an additional evaluation dimension \cite{kapoor2025agents,xia2024chatrepair,bai2026tokens}, or prescribe which agents may communicate with which \cite{qian2024chatdev,hong2024metagpt,zhuge2024gptswarm}. These measures do not show the coordination structure that emerges during a run.

We make that structure measurable. Each run is represented as a temporal network in which agents and files are nodes. A direct message is an agent-to-agent edge, a file write is an agent-to-file edge, and a file read is a file-to-agent edge; every edge carries a timestamp, a byte size, and a token cost. Treating files as nodes is central to the instrument. A file persists after it is written; one write can be read by many agents in any order; and its contents remain in the shared workspace across agent processes and outlive the session that wrote them. Files therefore provide persistent shared state as well as a one-to-many communication channel. Representing them as nodes allows file-mediated coordination and direct messaging to be measured on the same timeline. Every edge corresponds to a logged event, so the temporal network records the coordination that occurred during the run.

We apply the instrument to 1{,}902 runs, each evaluated with a fixed test suite, across two experiments using synthetic Python tasks. In Experiment~1, the \emph{distributed} task, each agent holds a different part of a specification and the team must reconstruct the whole. In Experiment~2, the \emph{chained} task, each agent owns consecutive steps of a processing chain and the team must agree the interfaces between them. Across these runs we vary three factors: the team size, from one to eight agents in the main experiments and up to sixteen in a scaling arm\footnote{A \emph{scaling arm} is a pre-registered extension of the chained task with more steps, so that a larger team still has work for every agent.}; the team structure, with either a flat team or one agent named coordinator; and the file policy, with shared files forbidden, allowed, or mandatory. Section~\ref{sec:experiment} defines the experimental design in full.

One question organises the study: \textbf{\emph{how does coordination scale and reorganise as these teams grow, under different team structures and file policies?}} The answer speaks to each of the three configuration choices above, and in each case the coordination we measure behaves differently from what the run's outputs alone would suggest.

\begin{itemize}
  \item \textbf{Team size.} Messaging does grow near-quadratically with the team, which suggests a cost that will keep growing. The timestamps show otherwise: the growth is a one-off round of introductions early in each run, what persists afterwards is far smaller, and on the largest teams messaging growth levels off as the teams shift towards broadcast (Section~\ref{sec:handshake}).
  \item \textbf{Communication channel.} A direct message reaches one agent; a file is written once and can be read by many. Requiring teams to coordinate through shared files cuts a large share of the output-token cost on one task and adds cost on the other. This difference shows that the effect of the channel depends on the task (Section~\ref{sec:files}).
  \item \textbf{Leadership.} Naming one agent the coordinator in its prompt produces no hub and no reliable gain in success, and a sealed replication finds flat and coordinator teams level at eight agents. A team is organised by the structure that emerges in its interactions, and a prompt label alone creates none of it (Sections~\ref{sec:leadership} and~\ref{sec:containment}).
\end{itemize}

Underneath all three choices is the shape of the network itself, and the task sets that shape. The distributed task builds a dense, tightly clustered mesh that rides the all-to-all line as the team grows; the chained task builds a sparse network whose gap to all-to-all widens with team size, until at sixteen agents scarcely any named-message network forms at all (mean degree 0.28 against a clique of fifteen). Neither shape has a leader. Left to organise themselves, the teams do not converge on a single coordination span; they let the task dictate the structure (Section~\ref{sec:topology}).

A further result concerns the instrument itself. Under a pinned model, the chained task reproduces across collection sessions almost exactly, while the distributed task does not: the same configuration, collected twice under the same pinned model, gave two incompatible growth rates. A configuration is therefore characterised by a distribution of runs, and any single run is one sample from it; where matched sessions exist we report the headline number with its cross-session range (the scaling arms and the sealed replication are single batches; Section~\ref{sec:reliability}).

The measurements also expose failures that the final code hides. An eight-step calculation split one step per agent failed in all ten runs, every time on the same question: round at each step, or once at the end. The convention sat on the boundary between two agents, and no agent owned it. The teams discussed rounding in every one of those runs and still never agreed. Success rates do not reveal this failure mechanism. The temporal network shows the structural condition behind it: the failing interface lies between two different owners (Section~\ref{sec:leadership}).

A last result is a behaviour we did not anticipate. Each task is graded by a hidden test suite, and no prompt tells the agents to look for it. Yet the teams went looking, and a validity check found them reading it, together with the reference solution, where the main runs had left both reachable. We re-ran the load-bearing cells in a sealed environment, with decoys in place of the hidden files: the two findings these cells test reproduce, and the teams still reach for the hidden test suite in four fifths of runs even though it returns nothing, an unprompted search for the answer key that a grade cannot see (Section~\ref{sec:containment}).

The paper contributes: (i) a temporal-network representation of multi-agent coding runs with agents and files as first-class nodes; (ii) an instrumentation pipeline that turns a run's logs into that network; (iii) a released dataset of 1{,}902 graded, fully instrumented runs across a controlled grid of configurations, together with a sealed replication of a further 244 runs (Section~\ref{sec:containment}); and (iv) the three findings above, each pre-registered where confirmatory and reported with its reliability, two of them (the coordinator null and the file-channel substitution) re-tested in the sealed replication, together with a task-resolved topology result showing that the task sets the shape of the coordination network (Section~\ref{sec:topology}).
\section{The Instrument: a Run as a Temporal Network}
\label{sec:instrument}

\begin{figure}[t]
  \centering
  % The principal methodology figure. One small run drawn as the measured
% object: agents and files are both nodes, every logged event is a typed
% edge, and the circled numbers give the order in time, so the "temporal"
% part of the graph is visible. spec.md is written once and read twice,
% quietly making the one-to-many point that Finding 2 develops.
% Layout is crossing-free by construction.
\begin{tikzpicture}[
  agent/.style = {circle, draw=black!75, fill=blue!8,
                  minimum size=8.5mm, font=\small, inner sep=0pt},
  file/.style  = {rectangle, draw=black!75, fill=orange!14,
                  rounded corners=2pt, minimum width=17mm, minimum height=7.5mm,
                  font=\small\ttfamily, inner sep=3pt},
  msg/.style   = {-{Stealth[length=2.4mm]}, thick, draw=black!85},
  write/.style = {-{Stealth[length=2.4mm]}, thick, draw=blue!65!black, dashed},
  read/.style  = {-{Stealth[length=2.4mm]}, thick, draw=orange!80!black,
                  dash pattern=on 0.6pt off 2.2pt, line cap=round},
  ev/.style    = {circle, solid, draw=black!60, fill=white, inner sep=0.6pt,
                  font=\tiny, minimum size=3.4mm}
]
  % ---- nodes -------------------------------------------------------------
  \node[agent] (a1) at (0,2.4)   {$a_1$};
  \node[agent] (a2) at (2.6,2.4) {$a_2$};
  \node[agent] (a3) at (0,0)     {$a_3$};
  \node[agent] (a4) at (2.6,0)   {$a_4$};

  \node[file] (spec) at (6.0,2.4) {spec.md};
  \node[file] (sol)  at (6.0,0)   {solution.py};

  % ---- messages (black, solid) --------------------------------------------
  \draw[msg] (a1) -- (a2) node[ev, midway] {1};
  \draw[msg] (a1) -- (a3) node[ev, midway] {2};
  \draw[msg] ([yshift=2.5pt]a3.east) -- ([yshift=2.5pt]a4.west) node[ev, midway, above=0.5pt] {6};
  \draw[msg] ([yshift=-2.5pt]a4.west) -- ([yshift=-2.5pt]a3.east) node[ev, midway, below=0.5pt] {7};

  % ---- file writes (blue, dashed) ------------------------------------------
  \draw[write] (a2) -- (spec) node[ev, midway] {3};
  \draw[write] (a4) -- (sol)  node[ev, midway] {8};

  % ---- file reads (orange, dotted) -----------------------------------------
  \draw[read] (spec.south west) -- (a3.north east) node[ev, pos=0.45] {4};
  \draw[read] (spec.south) to[bend left=12] node[ev, pos=0.5] {5} (a4.north east);

  % ---- legend ---------------------------------------------------------------
  \begin{scope}[shift={(-0.4,-1.45)}, font=\footnotesize]
    \draw[msg]   (0,0) -- (0.85,0);      \node[anchor=west] at (0.9,0)  {message};
    \draw[write] (2.65,0) -- (3.5,0);    \node[anchor=west] at (3.55,0) {file write};
    \draw[read]  (5.35,0) -- (6.2,0);    \node[anchor=west] at (6.25,0) {file read};
    \node[ev] at (7.95,0) {1};
    \node[anchor=west] at (8.15,0) {order in time};
  \end{scope}
\end{tikzpicture}
  \caption{One small run drawn as the measured object. Agents (circles) and files (rectangles) are both nodes; every logged event is a typed edge carrying a timestamp, a byte size, and a token cost, and the circled numbers give the order in time. Here $a_1$ opens with two messages (1, 2), $a_2$ writes the specification once (3), two agents read it (4, 5), $a_3$ and $a_4$ agree an interface by message (6, 7), and $a_4$ writes the deliverable (8). What to notice: the files take part in the network as nodes of their own, and one write (3) serves two readers (4, 5); a direct message reaches one recipient.}
  \label{fig:graph}
\end{figure}

Formally, we model each run's temporal network as a heterogeneous graph $G=(V,E)$. The vertices $V = A \cup F$ are the agents $A$ and the files $F$ touched during the run. The edges are typed and timestamped: an agent-to-agent edge is a direct message, an agent-to-file edge is a write or an edit, and a file-to-agent edge is a read. Every edge carries a timestamp, a byte size, and a token cost. Figure~\ref{fig:graph} shows a small instance.

Making files nodes, with their own identity and their own edges, is the design choice everything else rests on. The file channel behaves unlike the message channel in three ways. A file persists after it is written. One write can be read by many agents, in any order and after any delay. And a file crosses process and credential boundaries that an in-memory channel cannot. Giving files separate read and write edges puts file-mediated coordination on the same footing as direct messaging, which is what makes the channel comparison of Section~\ref{sec:files} possible at all.

The network is a direct record: every edge corresponds to a logged tool-call event, a message sent through a dedicated messaging tool or a file operation performed with the runtime's file tools, so the counts and identities of those events are exact. One blind spot follows from this: a file read or write issued through the shell leaves no edge, so file activity is a slight undercount (Section~\ref{sec:threats}). The one coarse quantity is the token cost on file edges: the runtime reports tokens per turn, and a turn can make several tool calls, so we divide a turn's output tokens evenly across its calls. The authoritative per-turn usage is stored alongside, so any finer attribution can be applied later without re-parsing.

\begin{figure}[t]
  \centering
  % The experimental run pipeline. Two columns: the run-time main flow on
% the left, the instrumentation and parsing on the right. Master CSVs
% (cross-run aggregation) and the resumable ledger are described in the
% caption and the surrounding prose, not drawn here, to keep the figure
% free of crossing arrows.
\begin{tikzpicture}[
  font=\footnotesize,
  node distance = 4mm and 6mm,
  box/.style    = {rectangle, draw=black!75, rounded corners=1pt,
                   minimum width=26mm, minimum height=8mm, align=center,
                   inner sep=2pt, fill=blue!4},
  store/.style  = {rectangle, draw=black!75, rounded corners=1pt,
                   minimum width=26mm, minimum height=8mm, align=center,
                   inner sep=2pt, fill=orange!8, font=\footnotesize\ttfamily},
  side/.style   = {rectangle, draw=black!75, rounded corners=1pt,
                   minimum width=26mm, minimum height=8mm, align=center,
                   inner sep=2pt, fill=gray!8},
  flow/.style   = {-{Stealth[length=2mm]}, semithick, draw=black!85},
  twoflow/.style= {{Stealth[length=2mm]}-{Stealth[length=2mm]}, semithick,
                   draw=black!85}
]

  % Left column: run-time main pipeline, top to bottom.
  \node[box]   (lib)    {Task library};
  \node[box]   (gen)    [below=of lib]    {Task generator\\\scriptsize(one configuration)};
  \node[box]   (setup)  [below=of gen]    {Run setup\\\scriptsize(workspace, prompts)};
  \node[box]   (agents) [below=of setup]  {$N$ agent processes\\\scriptsize(parallel)};
  \node[box]   (verif)  [below=of agents] {Verifier on workspace};
  \node[store] (out)    [below=of verif]  {result.json};

  % Right column: instrumentation and parsing, top to bottom. Anchored as a
  % single vertical stack off the MCP node (itself aligned to the agents row),
  % so the column shares one x-position and the arrows stay vertical. Box
  % heights and row spacing match the left column, so the rows still align.
  \node[side]  (mcp)    [right=of agents] {MCP server\\\scriptsize(message log)};
  \node[store] (logs)   [below=of mcp]    {session +\\message logs};
  \node[box]   (parse)  [below=of logs]   {Parser};
  \node[store] (csvs)   [below=of parse]  {per-run CSVs:\\\scriptsize nodes, edges, turns, runs};

  % Main pipeline arrows (pure vertical, left column).
  \draw[flow] (lib)    -- (gen);
  \draw[flow] (gen)    -- (setup);
  \draw[flow] (setup)  -- (agents);
  \draw[flow] (agents) -- (verif);
  \draw[flow] (verif)  -- (out);

  % Agents <-> MCP server (pure horizontal).
  \draw[twoflow] (agents) -- (mcp);

  % Instrumentation pipeline (pure vertical, right column).
  \draw[flow] (mcp)    -- (logs);
  \draw[flow] (logs)   -- (parse);
  \draw[flow] (parse)  -- (csvs);

\end{tikzpicture}
  \caption{From a run to a network. The generator deals the task's units to the agents, the agents run as parallel processes and message each other through a logging server, a verifier grades the workspace, and a parser turns the logs into four CSV tables: nodes, edges, turns, and runs. What to notice: instrumentation is external to the agents, so the pipeline works for any runtime that logs tool calls.}
  \label{fig:pipeline}
\end{figure}
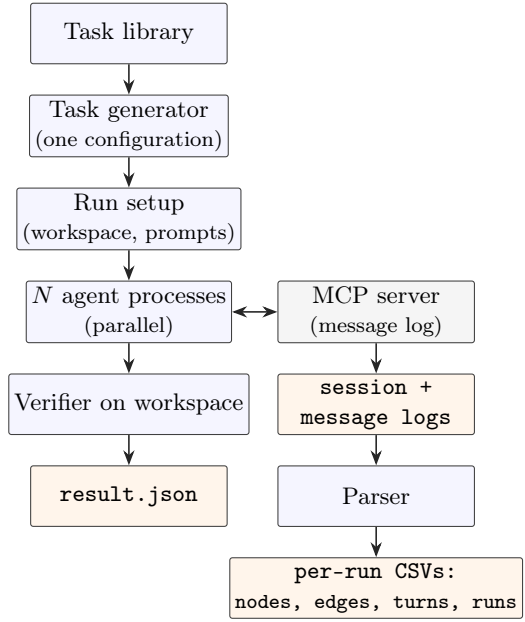

Figure~\ref{fig:pipeline} shows the pipeline. A generator produces a task instance for the requested configuration and deals its units out to the $N$ agents. Each agent runs as a separate process, sends direct messages with a \texttt{send\_message} tool, and collects them with \texttt{check\_messages}; every message is logged with sender, recipient, size, and time. When the team finishes, a verifier runs a fixed test suite against the workspace, and a parser converts the session and message logs into four linked CSV tables that encode the network. One run yields one network and one binary outcome, and that pair is the unit of analysis.
\section{The Experiment}
\label{sec:experiment}

\begin{figure}[t]
  \centering
  \resizebox{\textwidth}{!}{% The dataset at a glance, drawn rather than written:
%   top    - pictograms of the two task shapes;
%   middle - the configuration space as a stacked grid, one cell magnified
%            into what a cell contains (10 runs -> 10 graded graphs);
%   bottom - the run totals.
\begin{tikzpicture}[
  font=\footnotesize,
  agent/.style = {circle, draw=black!75, fill=blue!10, minimum size=5.5mm,
                  inner sep=0pt, font=\scriptsize},
  piece/.style = {rectangle, draw=black!70, minimum width=11.5mm, minimum height=6.5mm,
                  inner sep=1pt, font=\tiny},
  stepbox/.style = {rectangle, draw=black!70, rounded corners=1pt, fill=orange!10,
                  minimum width=13.5mm, minimum height=6mm, inner sep=1pt,
                  font=\tiny\ttfamily},
  own/.style   = {rectangle, draw=black!50, dashed, rounded corners=3pt, inner sep=2.5pt},
  deal/.style  = {-{Stealth[length=1.8mm]}, semithick, draw=black!70},
  chain/.style = {-{Stealth[length=1.8mm]}, semithick, draw=black!85},
  zoom/.style  = {draw=black!35, thin},
  note/.style  = {font=\scriptsize, text=black!70, align=center}
]

% ======================= TOP: the two task shapes =========================
\node[font=\footnotesize\bfseries] at (-3.8,0.55) {Experiment 1: distributed knowledge};
\node[note] at (-3.8,0.08) {\texttt{process\_orders}, one function, spec cut into four parts};

% the spec, cut into four pieces
\node[piece, fill=red!14]    (p1) at (-5.6,-0.75) {signature};
\node[piece, fill=blue!14]   (p2) at (-4.4,-0.75) {validation};
\node[piece, fill=green!16]  (p3) at (-3.2,-0.75) {discount};
\node[piece, fill=yellow!25] (p4) at (-2.0,-0.75) {sorting};

% dealt to four agents
\node[agent] (a1) at (-5.6,-2.0) {$a_1$};
\node[agent] (a2) at (-4.4,-2.0) {$a_2$};
\node[agent] (a3) at (-3.2,-2.0) {$a_3$};
\node[agent] (a4) at (-2.0,-2.0) {$a_4$};
\draw[deal] (p1) -- (a1);
\draw[deal] (p2) -- (a2);
\draw[deal] (p3) -- (a3);
\draw[deal] (p4) -- (a4);
\node[note] at (-3.8,-2.85) {each agent sees one part, never the whole:\\the team must reassemble the spec by coordinating};

\node[font=\footnotesize\bfseries] at (3.8,0.55) {Experiment 2: sequential dependency};
\node[note] at (3.8,0.08) {\texttt{summarise\_transactions}, a four-step chain};

% the chain, owned in consecutive blocks
\node[stepbox] (s1) at (1.3,-1.5)  {parse};
\node[stepbox] (s2) at (2.8,-1.5)  {validate};
\node[stepbox] (s3) at (4.8,-1.5)  {aggregate};
\node[stepbox] (s4) at (6.3,-1.5)  {format};
\draw[chain] (s1) -- (s2);
\draw[chain] (s3) -- (s4);

\node[own, fit=(s1)(s2)] (o1) {};
\node[own, fit=(s3)(s4)] (o2) {};
\node[agent] (b1) at (2.05,-0.55) {$a_1$};
\node[agent] (b2) at (5.55,-0.55) {$a_2$};
\draw[black!50] (b1) -- (o1.north);
\draw[black!50] (b2) -- (o2.north);
% the interface between owners
\draw[{Stealth[length=1.6mm]}-{Stealth[length=1.6mm]}, draw=red!70!black, semithick]
  (o1.east) -- (o2.west);
\node[font=\tiny, text=red!70!black] at (3.8,-2.05) {the interface};
\node[note] at (3.8,-2.85) {each agent owns consecutive steps, without knowing its position:\\the interface between owners must be agreed};

% ================== MIDDLE: the configuration space =======================
% three stacked 4x3 grids = team structure; columns = team size; rows = files
\begin{scope}[shift={(-5.45,-6.55)}]
  % back layers (team structure)
  \draw[black!30, fill=gray!4]  (0.56,0.40) rectangle (3.16,2.35);
  \draw[black!45, fill=gray!2]  (0.28,0.20) rectangle (2.88,2.15);
  % front layer with grid
  \draw[black!70, fill=white] (0,0) rectangle (2.6,1.95);
  \foreach \x in {0.65,1.3,1.95} \draw[black!40] (\x,0) -- (\x,1.95);
  \foreach \y in {0.65,1.3}      \draw[black!40] (0,\y) -- (2.6,\y);
  % one highlighted cell: n=8, allowed
  \fill[green!30] (1.95,0.65) rectangle (2.6,1.3);
  \draw[black!70] (1.95,0.65) rectangle (2.6,1.3);
  % axis: team size (columns)
  \foreach \x/\l in {0.325/1, 0.975/2, 1.625/4, 2.275/8}
    \node[font=\scriptsize] at (\x,-0.25) {\l};
  \node[font=\scriptsize\bfseries] at (1.3,-0.62) {team size};
  % axis: shared files (rows)
  \node[font=\scriptsize, anchor=east] at (-0.08,0.325) {forbidden};
  \node[font=\scriptsize, anchor=east] at (-0.08,0.975) {allowed};
  \node[font=\scriptsize, anchor=east] at (-0.08,1.625) {mandatory};
  \node[font=\scriptsize\bfseries, rotate=90] at (-1.95,0.975) {shared files};
  % axis: team structure (depth)
  \draw[-{Stealth[length=1.8mm]}, black!60] (2.7,2.05) -- (3.3,2.5);
  \node[font=\scriptsize, anchor=west, align=left] at (2.62,2.78)
    {\textbf{team structure}: flat (collected twice, as collection A and B) or \emph{coordinator}};
  % Experiment-1 extra axis + degeneracy note
  \node[font=\scriptsize, text=black!70, align=left, anchor=west] at (-1.5,-1.3)
    {Each experiment repeats this whole grid for each \textbf{spec split} (clean, overlapping, conflicting);\\
     combinations that make no sense are dropped (e.g.\ mandatory files at one agent).};
\end{scope}

% the magnifier: what one cell is
\draw[zoom] (-2.85,-5.25) -- (1.0,-4.7);
\draw[zoom] (-2.85,-5.90) -- (1.0,-6.5);
\node[rectangle, draw=black!70, rounded corners=2pt, fill=green!6, inner sep=5pt,
      align=left, anchor=west] (cellbox) at (1.0,-5.6)
  {\textbf{one cell} $=$ one configuration, run \textbf{10 times}\\[2pt]
   \tikz{\foreach \i in {1,...,10} \node[circle, draw=black!60, fill=blue!12,
         inner sep=1.6pt] at (0.34*\i,0) {};}\\[2pt]
   every run $\rightarrow$ pass/fail from the test suite\\
   \phantom{every run} $+$ one temporal network};

% ======================== BOTTOM: the totals ==============================
\node[rectangle, draw=black!85, thick, rounded corners=2pt, fill=green!6,
      inner sep=5pt, align=center] at (0,-9.35)
  {Experiment 1: $85$ cells $\rightarrow 850$ runs \quad
   Experiment 2: $87$ cells $\rightarrow 870$ runs \quad
   scaling arms: $30 + 70$ runs (to $n{=}16$) \quad
   pilots, checks: $82$\\[2pt]
   $\Rightarrow$ \textbf{1{,}902 graded runs released}, model pinned to \texttt{claude-sonnet-4-6}};

\end{tikzpicture}}
  \caption{The dataset at a glance. \textbf{Top:} the two task shapes. In Experiment 1 the specification is cut into four parts and dealt out, so the team must reassemble knowledge; in Experiment 2 the task is a chain of steps owned by different agents, so the team must agree the interfaces between owners. \textbf{Middle:} every run sets three factors, team size, team structure, and the file policy; one choice of all factors is a \emph{configuration}, and ten runs of a configuration in one collection form a \emph{cell} (magnified). \textbf{Bottom:} the resulting run counts. What to notice: the dataset is one controlled grid, so any two cells differ only in the factors named here.}
  \label{fig:design}
\end{figure}

Figure~\ref{fig:design} summarises the experimental design. This section walks through it top to bottom and defines the terms used throughout the paper. The pre-committed hypotheses and the statistical machinery are in Appendix~\ref{app:stats}.

\paragraph{Two experiments.}
Work divides across a team in two canonical ways: into independent pieces of knowledge, or into dependent steps. We built one experiment for each, where an \emph{experiment} is a task together with its variants. In \textbf{Experiment 1}, the distributed task (\texttt{process\_orders}), the specification of one Python function, \texttt{process\_orders(orders, config)}, is cut into four parts: the function signature and return shape (a dictionary of the processed orders, their count, their summed total, and a count of rejected orders); the validation rules (an order is valid when its amount is a non-negative number and its quantity a positive integer); the discount calculation (a bulk discount when the quantity reaches a threshold, and a loyalty discount for listed customers, both applied to the order's amount alone and never multiplied by its quantity); and the sort order of the output. No agent in a team is given the whole specification, and none is told who holds what, so the team can only recover it by coordinating. How the parts are dealt out is the \emph{split}, with three values: \emph{clean}, each part held by exactly one agent; \emph{overlapping}, some parts duplicated across two agents; and \emph{conflicting}, where two agents hold versions of the validation rule that disagree on a single point (whether a quantity of zero is valid), the test suite silently enforces one version, and neither agent is told which. The conflicting split therefore plants a disagreement that the team must first notice and then resolve the right way. In \textbf{Experiment 2}, the chained task (\texttt{summarise\_transactions}), the task is a four-step processing chain, and each agent owns one or more consecutive steps. Every agent's prompt carries the same team-level knowledge, the name and signature of the pipeline function the team must deliver, together with its private piece: the full input-to-output contract of its own step and the name of the file it must produce. The chain's length, the agent's position in it, and the content of the other steps are withheld, and every step after the first names only the step immediately upstream. The interfaces between owners must therefore be agreed during the run, and the team must also decide who writes the pipeline file that assembles the chain. Experiment 2 crosses the same three splits (its conflicting split disagrees on whether an amount of zero is valid), and two longer chains appear in the scaling arms below.

\paragraph{Three crossed factors.}
Every run fixes three factors. \emph{Team size} is 1, 2, 4, or 8 agents. \emph{Team structure} is either flat, all agents equal, or \emph{coordinator}\footnote{Labelled \texttt{orchestrator} in the released dataset and the pre-registration records.}, where one agent's prompt, and only that agent's prompt, names it the coordinator. \emph{File policy}\footnote{Recorded as \texttt{artefact\_policy} in the released dataset and the pre-registration records.} governs coordination through shared files: \emph{forbidden}, only the deliverable itself may be written, so the file traffic that remains under this policy is only the deliverable being written and read back; \emph{allowed}, the team does as it pleases, which makes this the default condition; or \emph{mandatory}, all inter-agent state must pass through files. One choice of all factors, plus the split, is a \emph{configuration}. Crossing the factors and dropping the combinations that make no sense (a mandatory file policy with a single agent, for example) leaves 58 distinct configurations per experiment. A \emph{cell} is ten runs of one configuration in one collection, the unit at which we report results; the flat structure above one agent was collected twice (Section~\ref{sec:reliability}), so each of its 27 configurations contributes two cells, and the crossing gives 85 cells per experiment. Experiment 2's released grid carries two further single-configuration baseline cells, both at four agents, flat, allowed, giving 87: the eight-step chain (\texttt{compute\_invoices}) at a grid-interior configuration, and a variant of the main task (\texttt{summarise\_transactions\_v2}) whose output ordering depends on a constant defined in another, non-adjacent step's specification, so the value can only arrive through coordination. The two experiments' grids together contain 1{,}720 runs.

\paragraph{One deliberate duplication.}
The flat structure was collected twice, with byte-for-byte identical prompts, in two collection runs we call \emph{collection A} and \emph{collection B}\footnote{They appear in the released data under the topology labels \texttt{solo} and \texttt{peer}.}; above one agent the two are wired identically and differ only in when they were collected). The two collections overlap in calendar time, so the divergence reported in Section~\ref{sec:reliability} cannot be attributed to drift between distant sessions. The duplication is deliberate: it gives every flat cell an exact same-configuration twin, which is what lets Section~\ref{sec:reliability} measure how well the whole instrument reproduces. When a result uses only one of the two sessions, we say which.

\paragraph{Overstaffed cells.}
The grid holds each task fixed while the team grows, because a fixed task is what makes two cells comparable; the price is that an eight-agent team on a four-part task leaves four agents with no part of the specification to hold. We keep those cells because overstaffing is a normal state of real teams: projects routinely carry more people than the work needs, and the extra members keep joining the coordination while producing little. The overstaffed cells measure that cost directly, what adding agents does to the coordination network when the work cannot absorb them. The four agents without a part turn out to be the team's most active coordinators: in the eight-agent Experiment 1 runs they send 62\% of all messages, 13.7 per agent per run against 8.4 for the agents holding parts, because an agent that holds nothing must ask the others for everything. The one conclusion these cells cannot support is a scaling law, since above four agents growth in coordination cannot be separated from the team running out of work.

\paragraph{The two scaling arms.}
A \emph{scaling arm} is a chained task with more steps, so that a larger team still has work for every agent. There are two arms, both pre-registered: for each, we wrote down what we expected to find and exactly how we would test it before any runs were collected, and the released package contains those records. The eight-step chain (\texttt{compute\_invoices}) runs at 2, 4, and 8 agents under the allowed policy (ten runs per cell, 30 runs); the sixteen-step chain (\texttt{process\_billing}) runs at 4, 8, and 16 agents, the only part of the study that reaches sixteen agents. Figure~\ref{fig:scaling} and hypothesis H7 use the arm's own cells; the grid's baseline \texttt{compute\_invoices} cell is separate. The sixteen-step arm carries \emph{twenty} runs per cell under the allowed policy instead of ten, decided in advance: more runs per cell sharpen the test of whether messaging growth stops at large team sizes (hypothesis H8, Appendix~\ref{app:stats}). A further ten-run mandatory cell at sixteen agents brings the arm to 70 runs. Steps are dealt in contiguous blocks, so along an arm each agent holds four consecutive steps, then two, then one as the team doubles. At the top of each arm every agent owns exactly one step, so nobody is idle and any slowing of coordination growth cannot be the team running out of work. The sixteen-step arm was collected as one interleaved batch over three days, with team sizes balanced within each day so the round-robin protects the size comparison, and it is not a cross-session replication. The remaining 82 runs are pilots (small trial batches run before the full schedules) and methodological checks, giving 1{,}902 runs in the main collection; a further 244 runs form the sealed replication of Section~\ref{sec:containment}.

\paragraph{Grading.}
Every task ships with a reference solution, and a fixed suite of behavioural input/output tests grades the team's deliverable against outputs fixed in advance\footnote{22 tests for Experiment 1's task, 25 for Experiment 2's main task; the longer chains carry their own suites.}. The tasks are pure functions with no randomness, and comparison is exact. The test suite is placed outside each agent's working directory; in the main collection that placement was not otherwise access-controlled, and Section~\ref{sec:containment} reports what the teams did with the access and re-runs the load-bearing cells under a seal that removes it. A run succeeds only if every test passes, so success is unambiguous while the path to it varies.

\paragraph{Environment and release.}
All runs use Claude Code (the 2.1.x series) with the model pinned to \texttt{claude-sonnet-4-6} and logged, unchanged, on every turn. The dataset, the task generators, the pipeline, and every analysis script are released as a replication package.

\paragraph{Statistical approach.}
The unit of analysis is the run: one graph, one pass/fail outcome. Success rates carry Clopper--Pearson exact 95\% intervals \cite{clopper1934}; success contrasts use Fisher's exact test \cite{fisher1934smrw}, continuous contrasts the Mann--Whitney test \cite{mann1947}, and each set of related contrasts is corrected with the Benjamini--Hochberg procedure \cite{benjamini1995}, reported as $p_{\mathrm{BH}}$. Eight hypotheses (H1--H8) are stated in advance; six are pre-registered in both prediction and test (H1, H4--H8), while H2 and H3 carry qualifications set out in Appendix~\ref{app:stats}. The appendix lists all eight with their outcomes, and the findings cite them by number. At ten runs a single cell's success rate carries a margin of roughly thirty percentage points, so single-cell readings are directional and the pooled contrasts carry the inferential weight.
\section{Finding 1: the Quadratic Cost of Scale is Mostly a Handshake}
\label{sec:handshake}

\begin{table}[t]
  \centering
  \caption{Mean edges per run as the team grows (flat team, collection B, files allowed, clean split, Experiment 1; ten runs per row). Messages rise steeply and file writes barely move. The file-read count at eight agents (38.8) is not a coordination signal: most of these reads are outside the workspace, the agents opening the reference solution, the hidden tests, and other runs' files (Section~\ref{sec:containment}); genuine reads of the team's own workspace are about fifteen per run. Cells are drawn from collection B alone; Section~\ref{sec:reliability} reports how far the two collections differ.}
  \label{tab:scaling}
  \begin{tabular}{rcrrr}
    \toprule
    agents & success & messages & file writes & file reads \\
    \midrule
    2 & 10/10 & 6.1 & 3.0 & 2.5 \\
    4 & 10/10 & 28.5 & 4.5 & 3.3 \\
    8 & 10/10 & 71.3 & 6.9 & 38.8 \\
    \bottomrule
  \end{tabular}
\end{table}

\begin{figure}[t]
  \centering
  \includegraphics[width=0.85\textwidth]{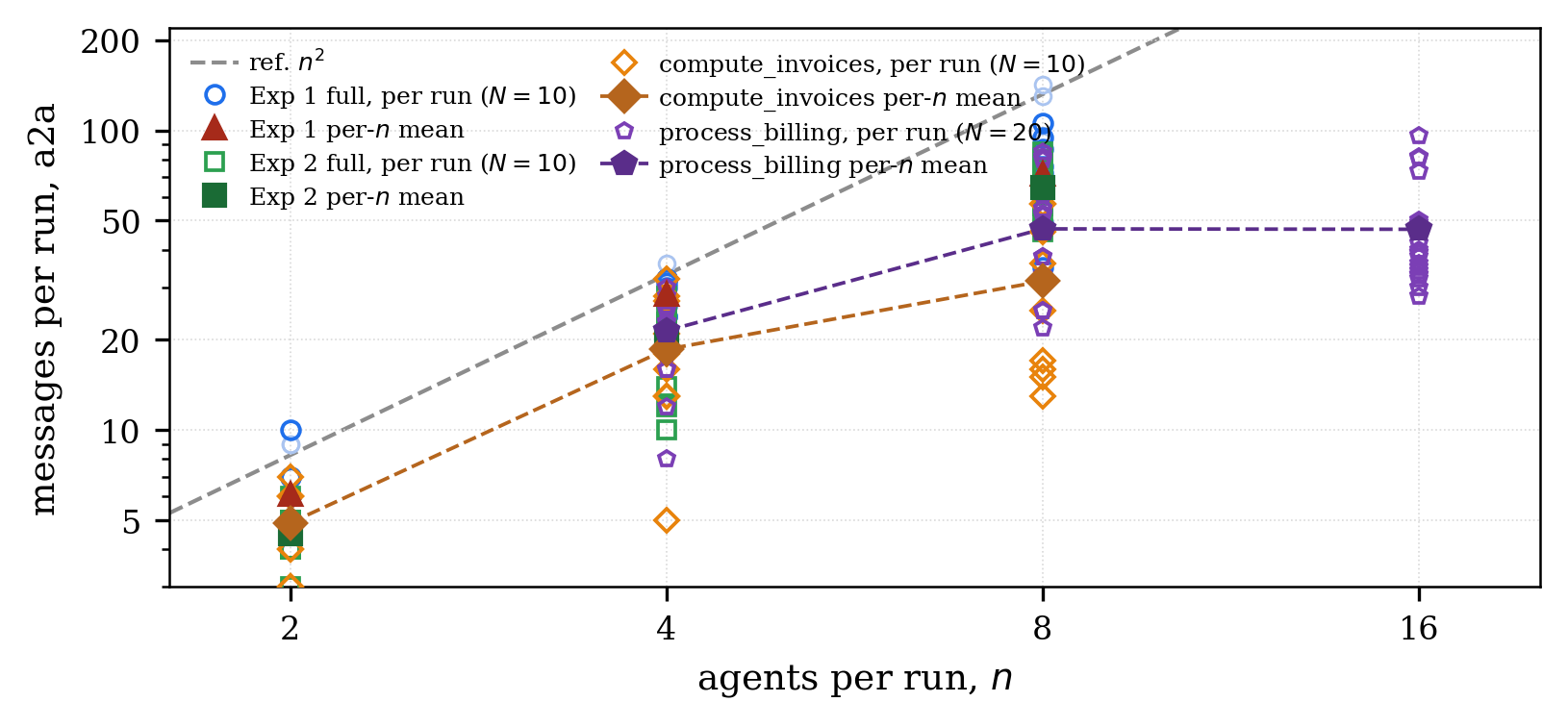}
  \caption{Messages per run against team size on log-log axes, with an $n^2$ reference (grey dashed); cells are drawn from collection B. What to notice: Experiment 2 tracks the quadratic reference (slope 1.92); the eight-step arm bends away from it after four agents; the sixteen-step arm goes flat between eight and sixteen. The quadratic growth ends within the measured range of team sizes.}
  \label{fig:scaling}
\end{figure}

If every agent must talk to every other, messaging scales as $n^2$ with team size, and the cost of coordination soon dominates the cost of the work. The first half of that picture holds in our data. Table~\ref{tab:scaling} shows messages per run rising from 6.1 at two agents to 71.3 at eight. Throughout the paper we quantify such growth by fitting a line to message count against team size on log-log axes; the slope of that line is the growth exponent, 1 for linear growth and 2 for quadratic. On the chained task the exponent is 1.92, quadratic within error, and the two collection sessions agree to two decimal places (1.92 and 1.93). The pre-registered H1 test, a regression over the three cell means, gives 1.92 with an interval of $[1.67, 2.17]$ ($\pm 1.96\,\mathrm{SE}$, the pre-registered test's normal approximation); the per-run regression we use for scaling elsewhere in the paper gives the same 1.92 with a tighter interval, $[1.80, 2.05]$. Both confirm quadratic scaling. On the distributed task the growth is also steep but not stable across sessions, a point Section~\ref{sec:reliability} takes up. Figure~\ref{fig:scaling} plots both experiments and the two scaling arms together.

The second half does not hold, and the timestamps show why. The growth is a round of introductions, made once, early. We read this from timing and message counts, because the instrument does not capture message content; \emph{introduction} here simply names a pair's first contact. Two measurements separate this opening handshake from the coordination that lasts.

First, the messages per pair fall as the team grows, from about three messages per ordered pair at two agents to 1.27 at eight. The total rises only because there are more pairs; each pair talks less.

\begin{figure}[t]
  \centering
  \includegraphics[width=0.9\textwidth]{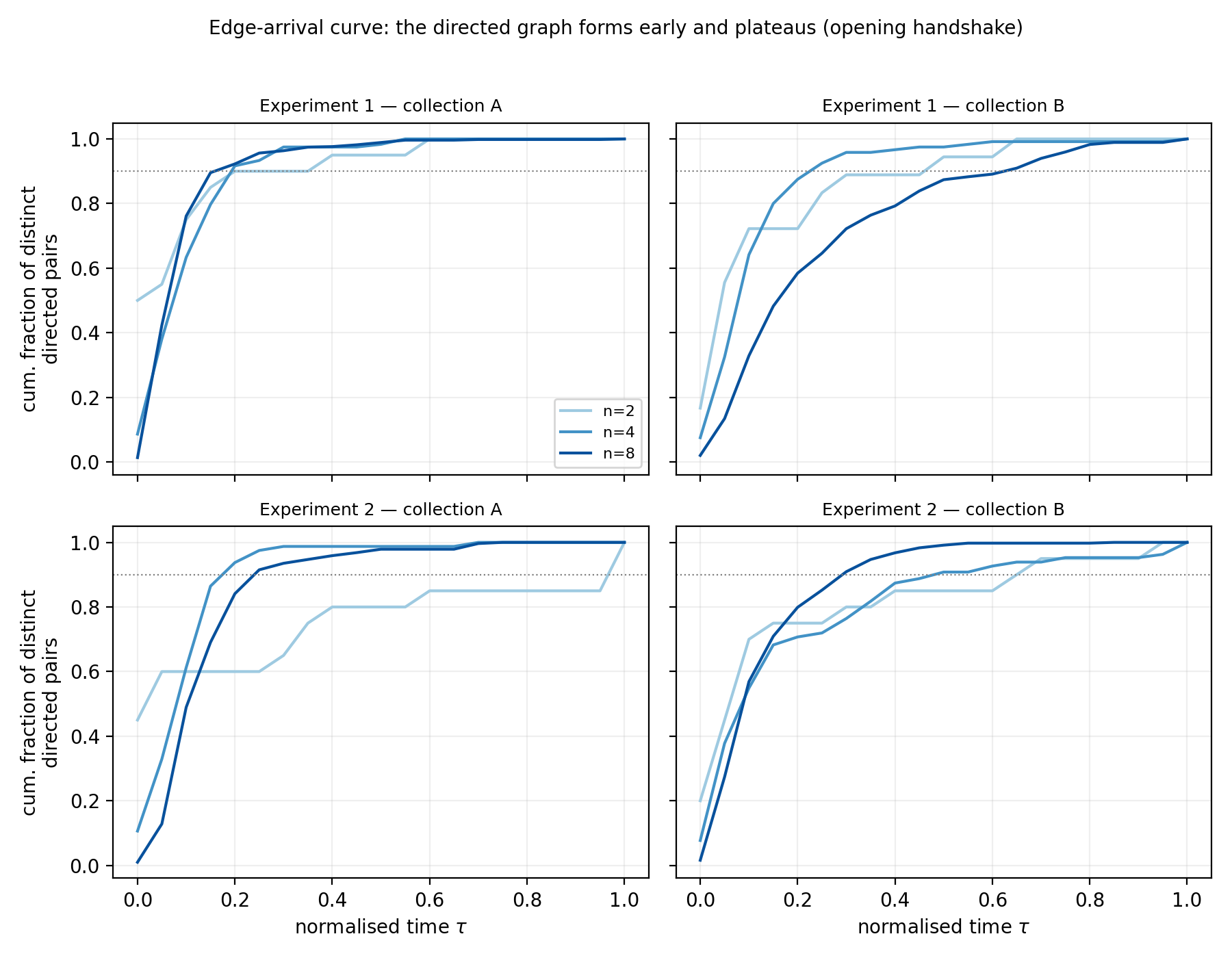}
  \caption{The handshake in time. Each curve averages, across timing-eligible runs that contain at least one named directed pair, the per-run cumulative fraction of the sender-to-recipient pairs a run eventually uses that have appeared by normalised time $\tau$; one line per team size, one panel per experiment and collection session, the dotted line marking 90\%. What to notice: most pairs appear early, though the mean per-run 90\% arrival time varies from about 0.14 to 0.60 across team sizes and sessions, the session dependence Section~\ref{sec:reliability} measures. The axis runs from the first message to the last, so it says nothing about where the run itself ends.}
  \label{fig:handshake}
\end{figure}

Second, the introductions come early in the run; the sustained traffic comes later. We place each run's messages on a normalised timeline, with $\tau = 0$ the run's first message and $\tau = 1$ its last, and report the mean over runs of the time by which ninety per cent of a run's pairs have appeared. On the distributed task, in one collection session, ninety per cent of the distinct sender-to-recipient pairs a run will ever use have appeared by $\tau \approx 0.2$ at every team size (Figure~\ref{fig:handshake}); in the other session the eight-agent handshake stretches to $\tau \approx 0.6$, so how early it completes is itself session-dependent. The chained task completes its handshake early at four and eight agents in both sessions; at two agents, in one session, ninety per cent of pairs appear only around the middle of the run ($\tau \approx 0.46$). What holds everywhere is the ordering, introductions first, then repeat traffic on a small established core, with the sustained channels running through the middle and second half of the run. Messages also get shorter as teams grow (mean size falls at every step in team size), so per-channel coordination thins on two axes at once.

The scaling arms close the argument, because each was collected as one interleaved batch with no idle agents and a decision rule fixed in advance. On the eight-step chain the slope falls from 1.82 between two and four agents to 0.72 between four and eight; at ten runs per cell the fall is in the predicted direction but not significant (hypothesis H7, Appendix~\ref{app:stats}). On the sixteen-step chain the growth stops outright: mean messages are 21.4, 47.0, and 46.8 at four, eight, and sixteen agents, the slope between eight and sixteen is 0.00 (95\% CI $[-0.34, 0.34]$), and the break is significant ($\Delta = 1.08$, 95\% CI $[0.61, 1.55]$), confirming hypothesis H8. Because that chain is fully divided only at sixteen agents, the plateau cannot be the team running out of work. Instead the team changes how it addresses itself: past eight agents, messages aimed at one named peer fall from 34.6 to 12.2 per run while broadcasts, messages sent to the whole team at once, rise from 12.3 to 34.0, and twelve of the twenty allowed sixteen-agent runs coordinate by broadcast alone. A large team stops addressing peers one at a time and speaks to the room.

\paragraph{Summary.} Adding an agent adds one more introduction, while the lasting per-agent coordination barely grows, and on the largest teams the messaging cost of growth falls to zero because the team switches to broadcast. The task sets the shape of the lasting traffic (Section~\ref{sec:topology}). A budget or a topology designed for sustained all-to-all messaging provides for traffic the teams do not produce.
\section{The Task Shapes the Network}
\label{sec:topology}

The handshake of Section~\ref{sec:handshake} says how much messaging there is; it does not say what shape the messaging settles into. To read the shape we build, for each run, its \emph{sustained undirected graph}: two agents are joined by an edge when at least one direction between them carried two or more messages during the run, so an edge marks a channel the pair used more than once; a single greeting does not count. We then measure two properties of that graph and average them over a cell's runs. The first is the mean degree, how many live partners the average agent keeps, read against the clique line $n-1$ that all-to-all coordination would draw. The second is the global clustering coefficient (transitivity), the share of an agent's partner-pairs that are themselves joined; it lies near 1 when the live partners form one tightly interconnected group and near 0 when they do not link to each other. Appendix~\ref{app:topology} gives the exact definitions and thresholds.

\begin{figure}[t]
  \centering
  \includegraphics[width=\textwidth]{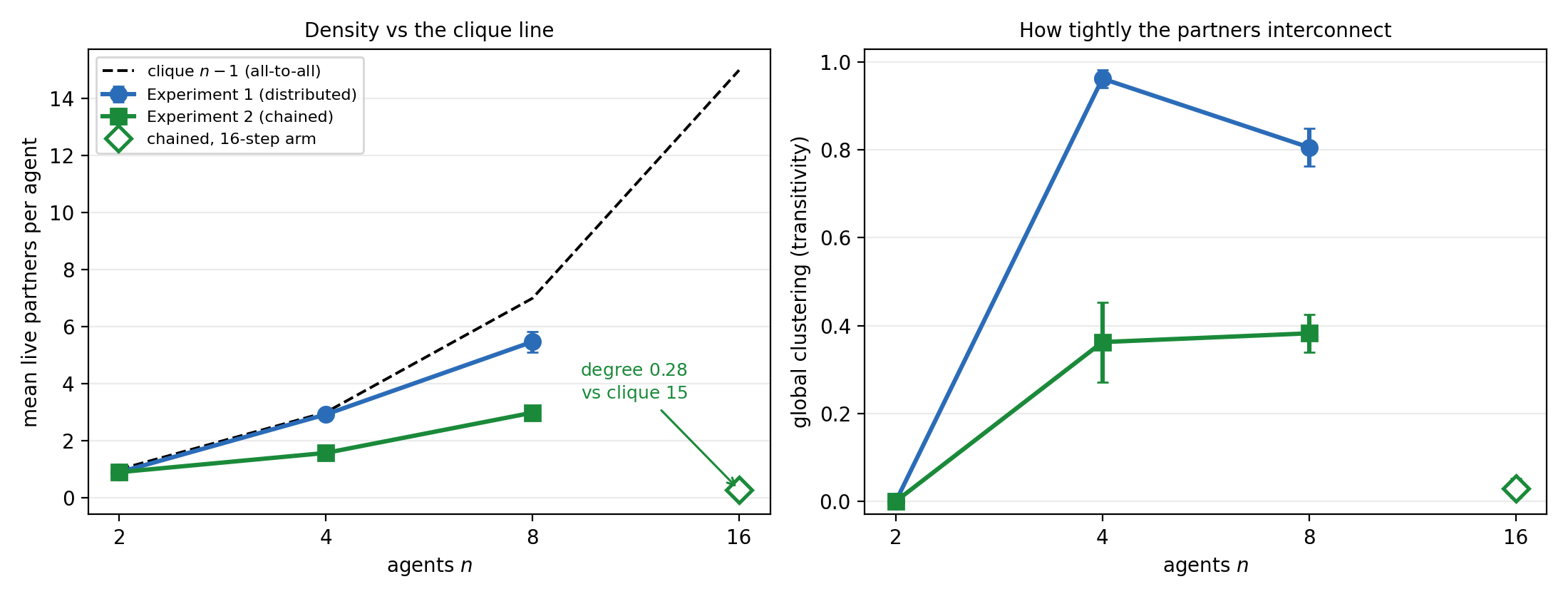}
  \caption{The shape of the sustained network is set by the task, on the flat, allowed-policy, clean-split runs (both collection sessions pooled). Left: mean degree of the sustained undirected graph (a pair is joined once either direction has carried at least two messages) against the clique line $n-1$ (black dashed), with standard-error bars; each two-, four-, or eight-agent point averages the two ten-run collection cells. Right: the global clustering coefficient of the same graph; at two agents it is zero by construction, since three agents are needed for a triangle. The open diamond is the chained sixteen-step scaling arm, the only sixteen-agent data. What to notice: the distributed task (blue) tracks the clique line and clusters tightly; the chained task (green) stays sparse and its gap to the clique line widens with team size; there are no sixteen-agent distributed runs.}
  \label{fig:topology-density}
\end{figure}

The shape is set by the task, and the two tasks settle on opposite shapes (Figure~\ref{fig:topology-density}). Both quantities are averaged over every configured agent and every run in a cell, so an idle agent counts as degree zero, and a run with no named-message network counts as zero and stays in the average; Appendix~\ref{app:topology} gives the estimator. On the distributed task, where every agent holds a fragment of one shared specification, the graph rides the clique line and fills in. Mean degree is 0.90, 2.92, and 5.47 at two, four, and eight agents, against clique lines of 1, 3, and 7, and clustering is high once the team is larger than a pair (0.96 at four agents, 0.81 at eight). Reconciling one specification pulls the team towards talking to everyone, and those partners also interconnect, so the sustained graph is a near-complete, tightly clustered mesh. On the chained task, where each agent owns consecutive steps of a pipeline, the graph stays sparse. Mean degree is 0.90, 1.57, and 2.99 at the same sizes, so the gap to the clique line widens as the team grows, and clustering stays low (0.36 at four agents, 0.38 at eight). A pipeline needs each owner to agree interfaces only with its neighbours, and the sustained graph records that local structure. The sixteen-step chain, the only configuration that reaches sixteen agents, has the widest gap: the average agent sustains just 0.28 partners against a clique line of 15, and clustering falls to 0.03. Only 8 of the 20 sixteen-agent runs carry any named agent-to-agent messaging at all; in the rest the named-peer network has gone quiet: the messaging that remains is broadcast, sent to the whole team at once (Section~\ref{sec:handshake}), and the coordination that is not in messages has moved to files (Section~\ref{sec:files}). Growing the team does not grow the named-peer network towards all-to-all; it stretches a sparse, chain-shaped graph thinner until, at sixteen agents, almost none of it forms.

Neither shape has a leader. The distributed mesh is dense but flat: no agent holds a disproportionate share of the sustained edges. Filtering each run's message graph to the channels that carry significantly more than an even spread (a disparity-filter backbone, $\alpha = 0.05$) leaves essentially no edges, 0 of 1{,}170 at eight agents on the distributed task and 2 of 1{,}077 on the chained task, and Section~\ref{sec:leadership} shows that naming a coordinator does not change this. The dense graph is dense everywhere and the sparse graph is sparse everywhere; neither concentrates on a hub. Whatever the task, the structure the team builds is leaderless.

\begin{figure}[t]
  \centering
  \includegraphics[width=0.85\textwidth]{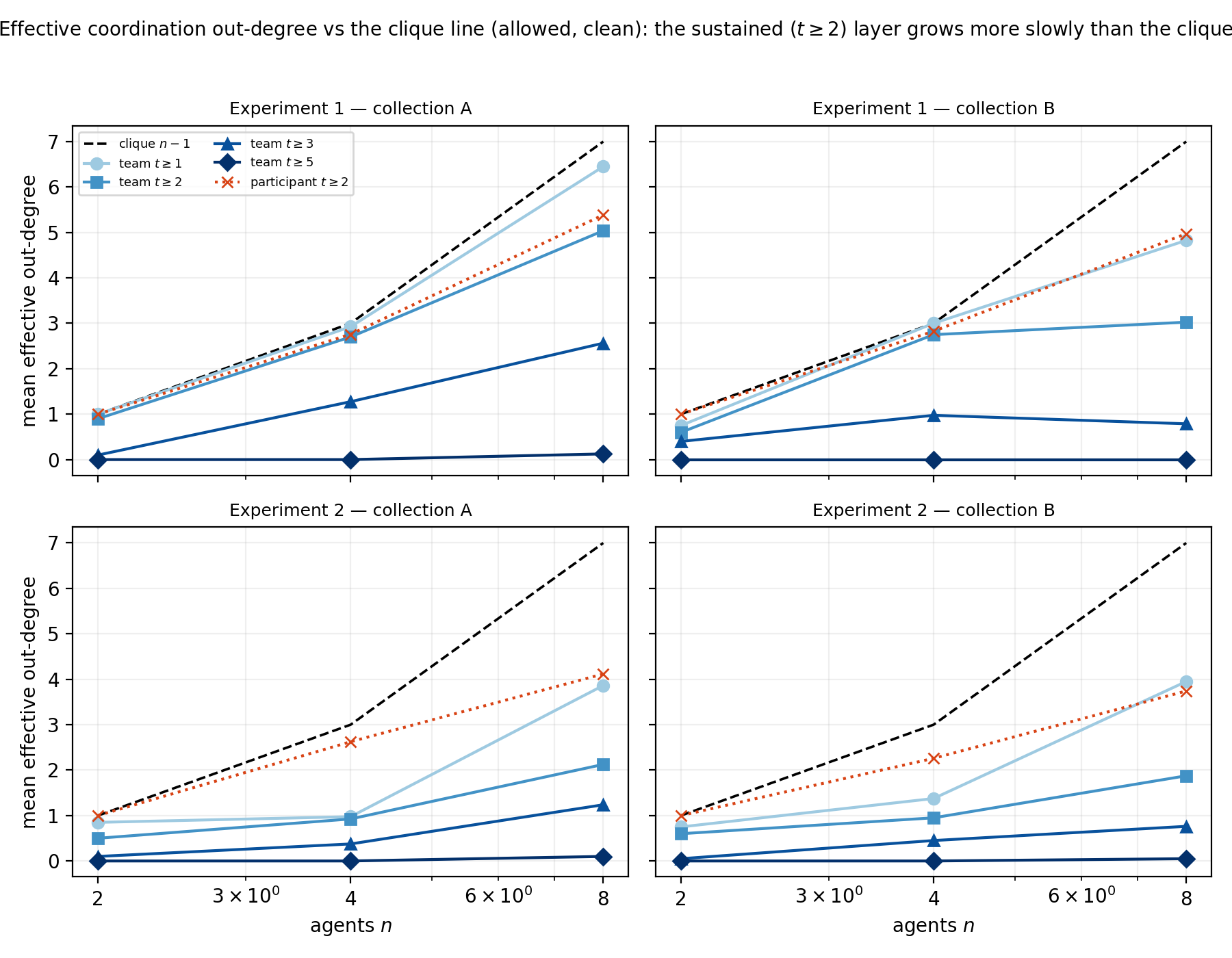}
  \caption{Mean directed out-degree against team size when edges are filtered to ordered pairs where the sender pushed at least $t$ messages, against the clique line $n-1$ (black dashed). What to notice: on the distributed task the once-only graph (light, $t \ge 1$) tracks the clique; the sustained directed graph ($t \ge 2$ and above) grows more slowly, to between about two and five per agent at eight agents depending on experiment and session (5.0 and 3.0 on the distributed task, about two on the chained). The participant series normalises over the agents that sent any message, the team series over all $N$ agents. This is an out-degree, and it sits below the dense \emph{undirected} graph of Figure~\ref{fig:topology-density} on the chained task and in one distributed session, though it approaches it in the other.}
  \label{fig:sustained}
\end{figure}

A directed version of this measure gives a narrower picture. If one counts, for the average agent, the number of distinct peers it \emph{sends} two or more messages to (a directed out-degree, normalised across all agents), the count is smaller and varies by experiment and session, roughly two to five at eight agents (Figure~\ref{fig:sustained}). It is a partial measure. On the chained task it stays around two; on the distributed task it is about three in one collection and five in the other, and in the second it comes close to the undirected degree of 5.5. The directed count describes how many peers an agent actively pushes to, which is generally fewer than the undirected connectivity, though it is a range, varying by experiment and session.

\paragraph{Summary.} Left to organise themselves, the teams do not converge on one coordination span. The distributed task builds a dense, tightly clustered mesh that rides the all-to-all line; the chained task builds a sparse graph whose gap to all-to-all widens with team size; and neither has a hub. The shape of the coordination network is a property of the task, read directly off the graph.
\section{Finding 2: Files are the Cheaper Channel, Where Messaging Would Dominate}
\label{sec:files}

The shape of the network is one half of the coordination cost; the channel it runs on is the other. A direct message reaches one recipient, so telling the whole team something means repeating it. A shared file is written once and read by many. The file policy of Section~\ref{sec:experiment} turns this difference into an experiment: the same tasks run with coordination files forbidden, allowed, or mandatory.

\begin{figure}[t]
  \centering
  \includegraphics[width=0.9\textwidth]{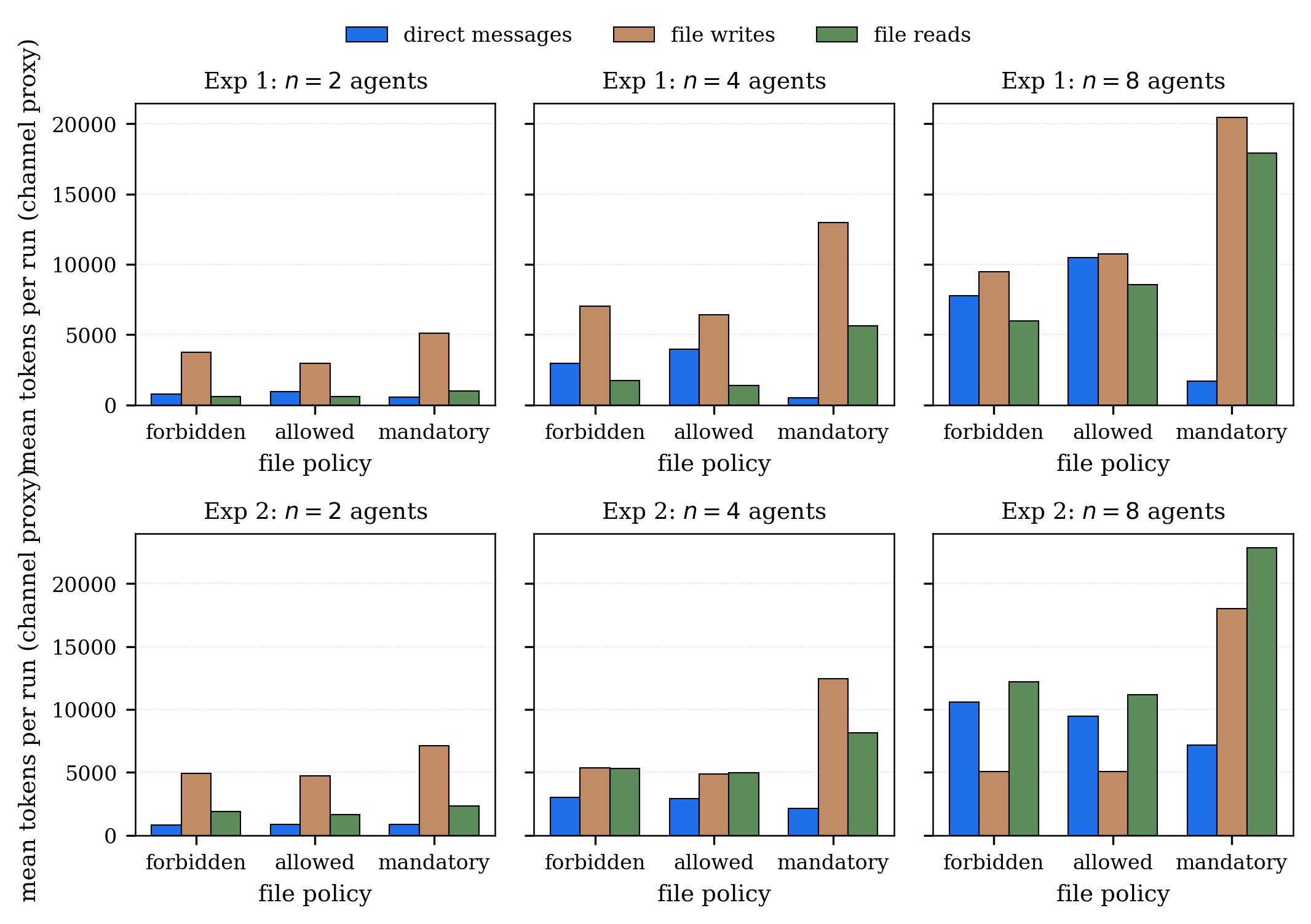}
  \caption{Where the coordination tokens go, by channel and policy (both experiments; 90 runs per bar group). The bars are proxies on two different bases: the message bar is the message length (about four characters per token), the file bars are the originating turn's output tokens shared across its tool calls, and neither counts cached context. They therefore show only the direction of the channel shift; the run-level output-token totals in the text use the model's reported figures. What to notice: under \emph{mandatory}, message tokens collapse and file tokens grow in both experiments. Output falls overall only in Experiment 1; in Experiment 2 the files already carry the coordination, so the rule only adds overhead. Under \emph{forbidden} the file bars are the deliverable itself, which the team must write: every write in those runs goes to the deliverable (270 of 270 multi-agent runs in Experiment 1), and the bars are non-zero because the same file is created and revised several times per run.}
  \label{fig:tokens}
\end{figure}

Mandating files changes where the coordination happens. At eight agents on the distributed task, the tokens attributed to direct messages fall from about 10{,}500 per run under the allowed default to 1{,}700 under mandatory, while file-write and file-read tokens roughly double (Figure~\ref{fig:tokens}). The two figures are measured differently, the message tokens from message length and the file tokens as a share of the turn's output (see the caption), so the decomposition shows only the direction of the shift; the direction is corroborated by the message count itself, which collapses under the policy, and by the run-level totals below. On the distributed task under the mandatory policy the distinct files a run touches grow with the team, from 3.2 at two agents to 11.9 at eight, and the pre-committed main effect, that mandatory adds file coordination while forbidden and allowed look alike, holds in both experiments (hypothesis H4).

The saving, however, belongs to one task shape, and we state it in the model's output tokens, which the run's logs report directly. On the distributed task the mandatory policy cuts output tokens by about 25\% at four agents and about 42\% at eight below the allowed default, with all team structures and splits pooled at each size (the eight-agent cut spans 36\% to 49\% across the two collection sessions). Output is only part of what a run consumes: most of the token throughput is cached context re-read on every turn, about 10.5 million tokens per run at eight agents under the allowed policy over the flat cells, which the run totals do not price, and the runs were collected on a subscription, with no per-token metering. The saving is not an artefact of the output-only view, since mandating files cuts the cached throughput as well, to about 6.6 million tokens per run. At API list rates for output alone, the eight-agent cut corresponds to roughly \$4.3 versus \$2.5 per run, an illustration of the output saving. On the chained task the same rule \emph{raises} output tokens, by about 17\% at four agents and 10\% at eight. The direction of the difference is the finding, and it makes the rule conditional: Experiment 1 teams left to themselves coordinate by one-to-one messages, so replacing that channel with files removes repetition; Experiment 2 teams already pass their work through files, each step's output read by the next, so mandating more file traffic only adds overhead. The sixteen-agent arm marks the limit sharply: with files already carrying the coordination, the mandatory rule roughly doubles per-run tokens (578k against 333k) and more than triples file reads, for identical success (10 of 10 against 20 of 20).

\begin{figure}[t]
  \centering
  \includegraphics[width=0.9\textwidth]{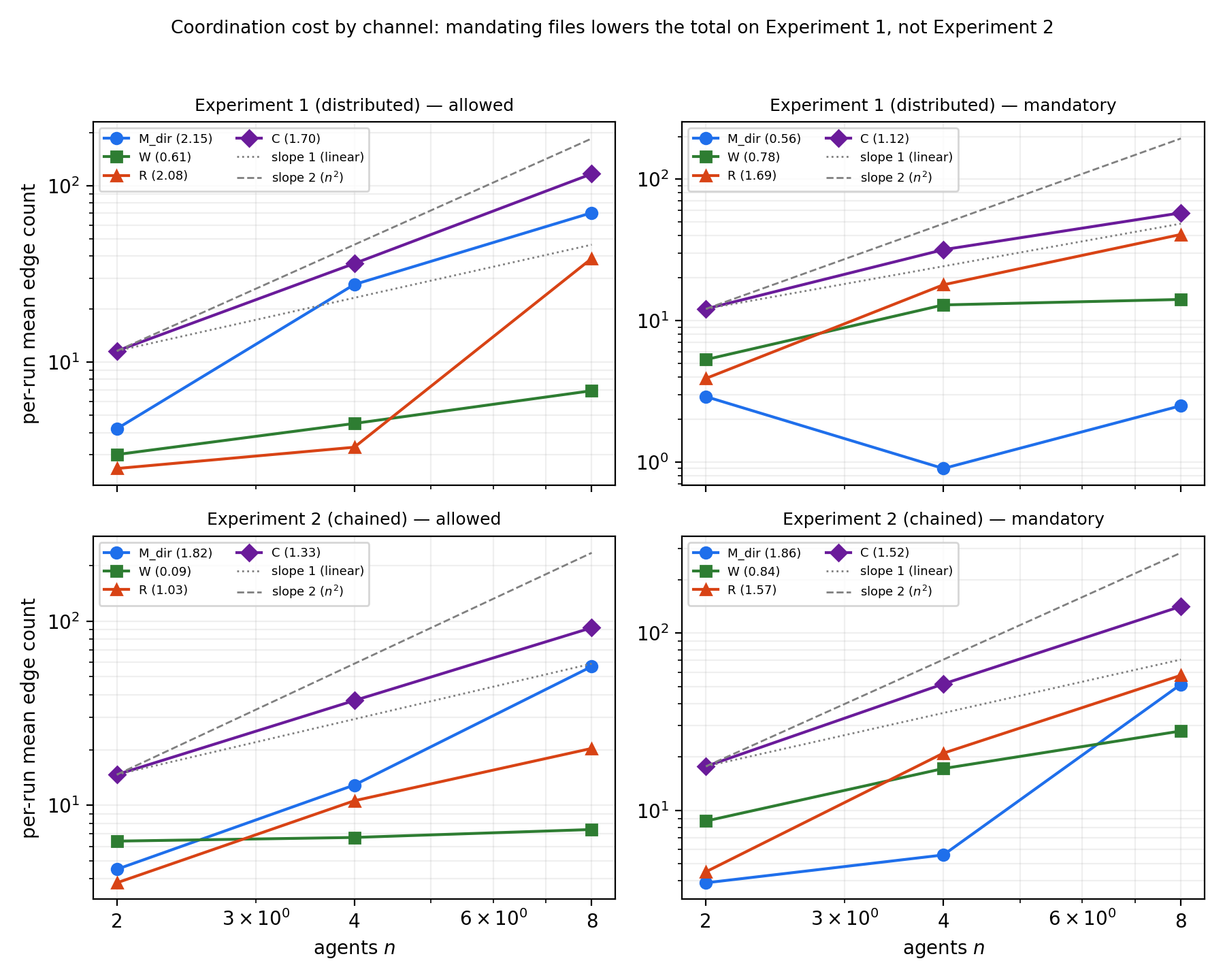}
  \caption{Coordination cost by channel against team size, log-log, under allowed (left) and mandatory (right), for the distributed task (top) and the chained task (bottom), all from collection B; the grey dashed line is $n^2$. Legend: M\_dir $=$ messages aimed at one named peer, W $=$ file writes, R $=$ file reads, C $=$ all coordination edges together; the number after each is its fitted growth exponent. What to notice: directed messages track the quadratic wherever the teams use them; on the distributed task the mandatory policy moves the total from near-quadratic (exponent 1.70 to 2.11 across sessions) towards linear (1.12 to 1.32). The file-read channel itself is still super-linear there ($\approx 1.7$): the total is near-linear because directed messaging is switched off. Only the chained task's file channel is linear by construction.}
  \label{fig:linearisation}
\end{figure}

The channel view explains the scaling as well as the totals (Figure~\ref{fig:linearisation}). The near-quadratic growth of Section~\ref{sec:handshake} belongs to the one-to-one message channel. Route the same coordination through shared files and the total-coordination exponent on the distributed task (all coordination edges, distinct from the messages-only exponent of Section~\ref{sec:reliability}) falls from 1.70--2.11 across sessions towards linear, 1.12--1.32; the cleanest reading, the eight-step chain at full decomposition where no agent is idle, gives $n^{0.98}$, linear within error. One caveat applies: on the distributed task the file-read channel alone still grows super-linearly (about 1.7), so the linearisation comes from removing the message channel, and the fully linear file regime is a property of chain-shaped work.

The policy also sets the run's shape in time. Messaging leads file writing (comparing each run's mean message time with its mean file-write time) in 86\% of allowed distributed-task runs, 62\% under forbidden, and 17\% under mandatory, where files must exist before anything can read them; a quarter of the mandatory runs (about a third at four and eight agents) send no message at all and drop out of this measure, so the 17\% describes the minority that still message. The chained task writes files first under every policy (messaging leads in only 12\% to 14\% of runs), even though its prompt asks agents to coordinate through messages. A sequential task settles its interfaces by writing them down, whatever it is told.

\paragraph{Summary.} Files are not passive outputs; they are the one-to-many channel, and the cheaper one wherever one-to-one messaging would otherwise dominate. The practical rule is conditional, and the graph tells you which side of it a task sits on: if the teams' own coordination is message-heavy, mandate files and save; if it already flows through files, leave it alone.
\section{Finding 3: Naming a Coordinator Does Not Create Structural Leadership}
\label{sec:leadership}

The simplest way to assign leadership to an agent team is a sentence in a prompt. In the coordinator condition one agent, and only that agent, is told it is the coordinator. If the label works, the communication graph should show it: traffic concentrating on the coordinator, and success rising on the tasks that need arbitration most.

Neither happens. No hub forms under either condition: filtering each run's message graph to the channels that carry a disproportionate share of an agent's traffic (a disparity-filter backbone, $\alpha = 0.05$) leaves 0 of 1{,}170 channels at eight agents on the distributed task and 2 of 1{,}077 on the chained task, and the same holds at four agents. A pre-committed null on the chained task also finds no difference in directed traffic at four agents in the pre-registered session (12.9 against 16.3 messages, $p = 0.29$). Whatever the prompt says, no communication hub emerges in either task. Nominal leadership does not become structural leadership.

\begin{table}[t]
  \centering
  \caption{Success on the conflicting split (Experiment 1), where two agents hold contradictory validation rules, by team size, policy, and structure; files-\emph{allowed} and the other policies shown separately, ten runs per cell. Collections A and B are the same flat configuration collected twice.}
  \label{tab:conflict}
  \begin{tabular}{rlccc}
    \toprule
    agents & policy & flat A & flat B & coordinator \\
    \midrule
    4 & forbidden & 8/10 & 8/10 & 7/10 \\
    4 & allowed   & 7/10 & 5/10 & 8/10 \\
    4 & mandatory & 7/10 & 5/10 & 8/10 \\
    8 & forbidden & 10/10 & 10/10 & 6/10 \\
    8 & allowed   & 8/10 & 10/10 & 8/10 \\
    8 & mandatory & 9/10 & 9/10 & 9/10 \\
    \bottomrule
  \end{tabular}
\end{table}

Success tells the same story once the two flat collections are pooled (Table~\ref{tab:conflict}). The pre-registered prediction was that a coordinator helps on the conflicting split, where two agents hold rules that disagree and someone must arbitrate. Success on this split measures arbitration only imperfectly: the grader silently enforces one of the two rules, so a team that arbitrates cleanly still fails if it agrees on the other, and the outcome mixes arbitration with landing on the enforced convention (the reference uses the stricter rule, that a zero quantity is invalid). At four agents the coordinator looks better (8/10 against the collection B's 5/10), but pooling the two flat collections dissolves the advantage (8/10 against 12/20, $p = 0.42$), and at eight agents the direction reverses: the flat team matches or beats the coordinator at every policy. The one contrast that survives correction across the six conflict cells is a reversal, flat 20/20 against coordinator 6/10 under forbidden ($p_{\mathrm{BH}} = 0.046$). This contrast does not reproduce once the grading suite is placed out of the agents' reach: the sealed replication of Section~\ref{sec:containment} runs the same eight-agent conflict cells and finds flat and coordinator teams level under every policy, so we treat the reversal as an artefact of the uncontained collection. On the chained task the coordinator with a free file channel is the weakest cell in the experiment (1/10). The pre-registered within-experiment comparison of the two topologies on the conflicting split is inconclusive (hypothesis H3); the cross-experiment interaction reported here ($p = 0.24$) is an exploratory test, outside the pre-registration (Appendix~\ref{app:stats}).

A flat team is no cure either. At two agents on the conflicting split, flat teams fail 37\% of their runs pooled across policies (up to half in the weaker session), while coordinating through the same channels as everyone else. Resolving a planted disagreement is not a matter of communication volume.

\begin{figure}[t]
  \centering
  \includegraphics[width=0.92\textwidth]{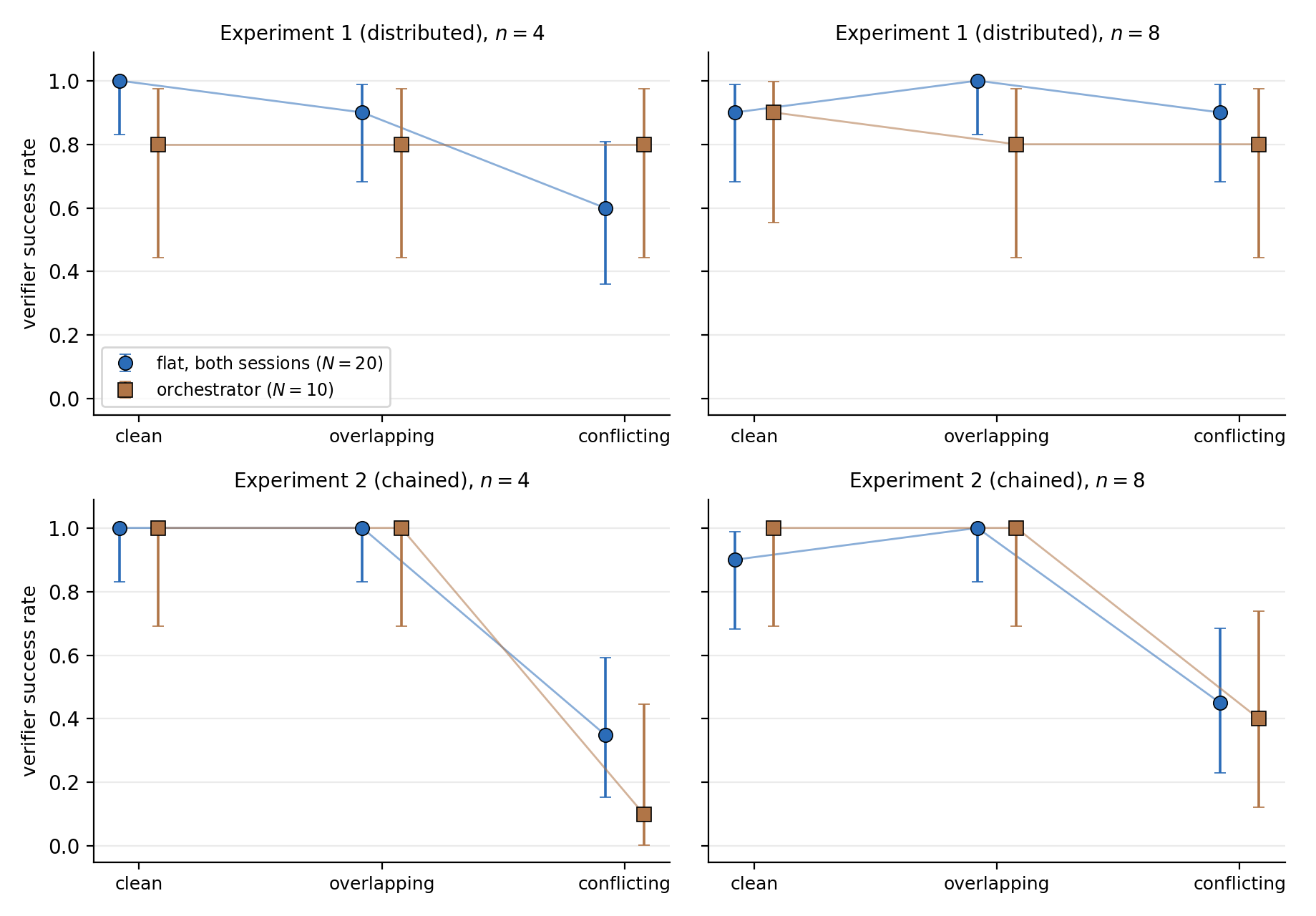}
  \caption{Success by team structure and split, under the allowed (default) file policy, with Clopper--Pearson 95\% intervals. Flat pools the two collection sessions ($N{=}20$ per point); coordinator has $N{=}10$. What to notice: at four agents the coordinator leads on the conflicting split (top left); at eight agents it no longer does (top right). In Experiment 2 (bottom) success follows the split, whatever the team structure, and the coordinator is the weakest point of all on the conflicting split at four agents (1/10).}
  \label{fig:topology}
\end{figure}

The sharpest failure has no leader and no fix by talking. The eight-step chain (\texttt{compute\_invoices}) succeeds in nine of ten runs at two and four agents, then fails all ten at eight agents, where each agent owns exactly one step. Every failure sits on the same seam, between \texttt{compute\_tax} (step 7) and \texttt{format\_invoices} (step 8): round at each step, or once at the end. The specification settles it, \texttt{compute\_tax} is told to pass its tax on unrounded and \texttt{format\_invoices} to round every figure once at the end, so the two agents each hold one half of the convention. At two and four agents one agent owns both steps and reconciles the halves internally; at eight agents the seam falls between two different owners and no agent holds responsibility for it. The teams were not silent: the run transcripts show rounding discussed in all ten of the eight-agent runs, and those teams messaged more than the four-agent ones. Talking more did not close an interface that nobody owned. The mismatch is invisible to each agent, whose own step is correct, and invisible in the finished code, which runs. Diagnosing it draws on the specification, the identical failure across the runs, and the transcripts, none of which is the graph; what the coordination graph adds is the structural signature, that at eight agents the seam falls on a boundary between two different owners, an interface no single agent sits across.

Beyond structure, no single network property buys success. Exploratory run-level regressions point one way, more messaging and writing go with failure and longer runs, more reading with success and shorter ones, but the associations explain about a tenth of the variance and the causality is open, since a run that goes well needs less repair traffic. The network's shape records how a run went more than it predicts how it will go.

\paragraph{Summary.} A coordinator exists only where the interaction structure carries the role, and a prompt clause alone does not create that structure. Decomposition creates interfaces between agents, every interface needs an owner, and the graph shows which interfaces have none; those are where the teams break.
\section{How Far to Trust One Run: Reliability of the Measurements}
\label{sec:reliability}

The flat condition was collected twice under byte-identical prompts, giving 54 matched cell pairs that are exact same-configuration replications under one pinned model. These replications quantify the reliability of the instrument. Where a task pins the coordination down the two sessions agree closely, showing that the instrument itself can reproduce consistently; where a task leaves the coordination open the spread between sessions is considerably larger and itself task-dependent, which uniform measurement noise would not produce.

\begin{figure}[t]
  \centering
  \includegraphics[width=0.62\textwidth]{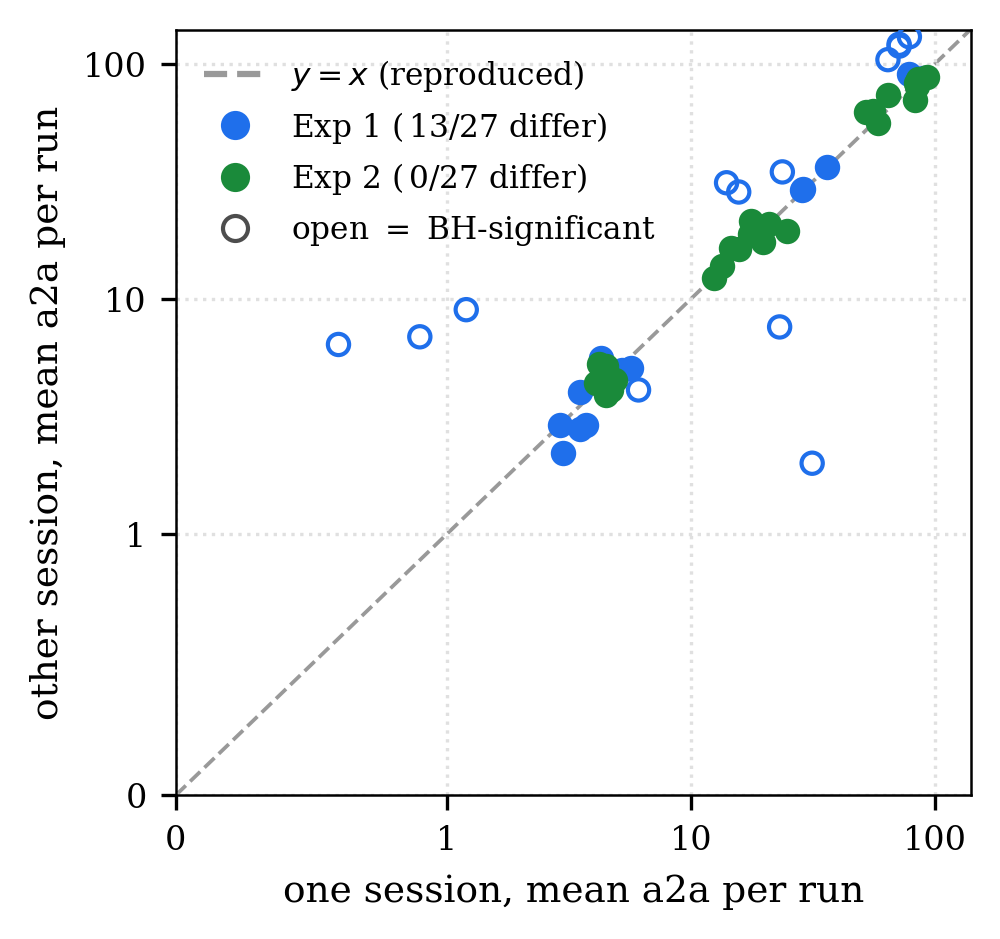}
  \caption{Test-retest reliability: each point is one matched cell, with the two sessions' mean messages per run on the two axes; open markers differ under a Benjamini--Hochberg correction across each task's cells. What to notice: the chained task (green) hugs the diagonal, with no cell differing after correction; the distributed task (blue) scatters off it, with thirteen of twenty-seven differing. Reproducibility under a fixed model depends on the task.}
  \label{fig:reliability}
\end{figure}

Reproducibility is task-dependent (Figure~\ref{fig:reliability}). On the chained task the sessions agree almost everywhere: no matched cell differs once the twenty-seven comparisons are Benjamini--Hochberg corrected (one differs at the raw level), and typical cell means move by about seven per cent. On the distributed task thirteen of twenty-seven cells differ after the same correction, one of them by a factor of fifteen in messaging (31.4 against 2.0 messages per run at the same configuration). File activity swings too: at eight agents under the allowed policy, one session's teams touch 18.0 distinct files per run on the conflicting split and the other's touch 4.3. This is why the distributed task's scaling exponent is a range, 1.76 (95\% CI $[1.57,1.96]$) in one session and 2.44 ($[2.32,2.57]$) in another, with non-overlapping intervals, and why the handshake completion time in Section~\ref{sec:handshake} is reported per session. The chained task's exponent, meanwhile, replicates to two decimal places.

The pattern in where the divergence lands explains it. The cells that drift are the high-coordination ones, where the task leaves the team the most freedom in how to organise; the cells that reproduce are those where the task pins the coordination down, as a chain does. The gap is not elapsed time between sessions, since the chained pairs span a longer wall-clock gap than the distributed ones and still reproduce.

\paragraph{Summary.} Under a fixed model identifier, a multi-agent configuration is characterised by a distribution of runs. Where the task leaves coordination open, single-run benchmarks of multi-agent systems measure a sample of size one from a wide distribution. Any instrument for these systems has to be used with repetition, and any headline number should carry its cross-session range where matched sessions exist, as ours do.
\section{Containment and What the Teams Reach For}
\label{sec:containment}

The runs analysed above were collected in an environment that did not confine the agents to their working directory. For each run the grading suite and the reference solution sat in sibling directories, and the run's manifest recorded their paths. A validity check made after the main collection found that the agents used this access: they opened the hidden test suite in 234 runs and read the reference solution in 77, the reach most common in the eight-agent conflicting cells. The same access reached further than the grading files: through the file tools, agents opened another agent's private prompt in 129 runs, read a different run's files in 18 (a finished solution in 16 of them), and read the shared message log in 52; shell-mediated reads, which the graph does not capture, would raise each of these counts. No prompt forbade reading these files, and none asked for them; the only stated constraint concerned which files a team could create. The agents went looking on their own initiative and reported what they found openly. The reachable files are a containment gap in the setup; the looking is the agents' own behaviour, measured directly under the seal below. The gap does not on its own decide the findings. The confirmatory results survive when every run with such a read is dropped, and the file-policy saving of Section~\ref{sec:files} is largest in the runs with no out-of-workspace read and smallest in the most-affected third when the affected runs are grouped by read count (46\% against 41\% across all affected runs and 37\% in the most-affected third), so the access dilutes that effect and cannot account for it. To settle the matter directly we re-ran the load-bearing cells in a sealed environment.

\paragraph{The seal.} Each run's working directory was placed on its own, with the grading suite, the reference solution, the manifest, and the other agents' prompts held outside it. Where the grading suite and the prompts used to sit we left decoys of the same names, each carrying a marked placeholder and none of the real content. Where the manifest used to sit we left a copy stripped of its grading pointers, the paths to the test suite and the reference solution; the copy still lists the team's component assignment, as the original did in the main collection, but nothing that grades the work. An agent that reached for the hidden material therefore found a file where it expected one, read a placeholder, and left a record of the attempt. The real files remained in place for the duration of each run for scoring, and each finished run's sensitive files were cleared before the next run began, but the relocation sends the reach the agents actually used, always a relative path out of the workspace, onto the decoy instead. Every read made through the file tools was logged and classified by location; we count the seal as holding only when no such read reached a real file, and we discarded and re-ran any run in which one did. In the event no run was discarded for such a read; four runs were discarded and re-run for agent timeouts. The seal therefore functions as a tripwire that defeats the observed reach pattern and records every tool-mediated attempt. It stops short of an enforced sandbox: a read issued through the shell would not be captured, and the runs execute under a permissive filesystem mode. The sealed batch is 244 runs at eight agents on the distributed task, covering the six conflicting-split cells (flat and coordinator under each file policy) and the two clean-split file cells, on the same pinned model and with the grading suite byte-identical to the released one. No tool-mediated read reached a real file in the runs we kept.

\paragraph{The teams reach for the answers.} The decoys turn the reach into a measurement. With the real material replaced by a placeholder and nothing to gain, the teams still went looking for it: in 80\% of the sealed runs an agent opened the decoy standing in for the hidden test file, in 66\% it opened another agent's prompt, and in 61\% it opened the run manifest. No prompt asks for any of these files; the test-file and prompt reads return only a placeholder, and the manifest read returns no grading material, so the reach for the test file is an unprompted tendency to seek out the grading material. It is the kind of reward-seeking that grade-only evaluation cannot see \cite{pan2022rewardmisspec,leike2017gridworlds,zhong2025impossiblebench,gabor2025evilgenie}, and it makes the containment gap in the main collection unsurprising in hindsight: agents that look for the answer key when they are confined will read it when they are not.

\paragraph{The findings hold under containment.} With the real material replaced by decoys, the two findings that these cells test reproduce. Naming a coordinator again makes no difference to success: flat and coordinator teams are level under every file policy (forbidden 20/30 against 21/32, $p = 1.00$; allowed 26/31 against 23/31, $p = 0.53$; mandatory 23/30 against 25/30, $p = 0.75$). The single reversal reported in Section~\ref{sec:leadership}, the flat team beating the coordinator under the forbidden policy, does not reappear, so it is best read as an artefact of the uncontained cells. The channel substitution of Section~\ref{sec:files} holds as well: mandating files collapses one-to-one messaging, from 108 messages per run under forbidden and 134 under allowed to 26 under mandatory on the conflicting split, and from 119 to 21 on the clean split, the same move from the message channel to the file channel measured in the main dataset.

\section{Threats to Validity}
\label{sec:threats}

\paragraph{Construct.}
The file graph captures tool-mediated file activity only. Edges come from the runtime's \texttt{Read}, \texttt{Write}, and \texttt{Edit} tools; a file operation issued through the shell leaves no edge. In the main collection this cost ten runs their recorded write, because their deliverable was created with a shell command, which leaves no edge, so file activity is undercounted in about 0.6\% of runs. This is an undercount, so the sparse file structure we report is conservative: any unlogged operation would only add edges. The per-channel token figures are proxies on two bases. Token cost on file edges divides a turn's output evenly across its tool calls, while token cost on message edges is estimated from message length, so the two are not a like-for-like total and the channel bars of Figure~\ref{fig:tokens} show direction only; the run-level output-token results use the model's reported totals and are independent of the division, and the direction of the channel substitution survives any monotone re-attribution within a turn. The reported token totals count input and output only, excluding the cached context re-read on each turn, which is the bulk of the throughput; the output saving of Section~\ref{sec:files} holds on the cached tokens too, and the runs were collected on a subscription, so we report token reductions. The coordinator is a prompt clause with no enforced routing, so Section~\ref{sec:leadership} measures the effect of nominal leadership; an enforced star topology is follow-on work, and the distinction is the point of the finding. The two experiments also differ in one prompt detail: Experiment 2's prompt asks agents to coordinate through messages, and its file-first behaviour runs against that instruction, so the cross-experiment contrasts are conservative with respect to it.

\paragraph{Internal.}
The agents are stochastic and a model identifier does not fix the provider's state. Every cell carries ten repetitions, the model identifier is logged per turn (no mid-study change occurred), and within a batch a regression of messaging on run order shows no drift. Between sessions the measurements do vary, and we analyse this variation explicitly in Section~\ref{sec:reliability} and report session ranges. At eight agents the clean main-matrix cells leave four agents idle, and the four-agent break coincides with the tasks' four-unit structure; the two scaling arms were built without either confound and reproduce the deceleration and the plateau. The main collection also did not confine the agents to their workspace, and Section~\ref{sec:containment} quantifies that access, shows the confirmatory findings survive the exclusion of every affected run, and re-runs the load-bearing cells under a seal that removes it.

\paragraph{External.}
The numbers come from two synthetic Python tasks, one runtime, and one pinned model, and the specific values will not carry directly to other settings. The claims we generalise are structural: the handshake-then-core shape, the one-to-many economics of files, and the gap between nominal and structural leadership. The graph schema and parser apply to any runtime that logs tool calls and messages; tasks from real repositories and other models are the natural replication.

\paragraph{Conclusion.}
Confirmatory claims rest on hypotheses pre-registered in both prediction and test (H1, H4--H8), with Benjamini--Hochberg correction within each contrast set; H2 and H3 carry the qualifications set out in Appendix~\ref{app:stats}, and the appendix also records that the pre-registered top-up rule flagged cells we did not top up. Power is the limit: at ten runs a cell's success rate carries a margin of roughly thirty percentage points, so single-cell readings are directional and pooled contrasts carry the weight. H7 met its pre-committed power-limited branch and is reported as directional; H8, at twenty runs per cell, reached significance.

\section{Related Work}
\label{sec:related}
Work on multi-agent LLM systems evaluates them, almost without exception, by their outputs. Benchmark studies score end-task success across frameworks and team configurations: MultiAgentBench evaluates different coordination topologies and, in an ablation study, different team sizes, and measures task completion and coordination performance \cite{zhu2025multiagentbench}; software-engineering studies compare role-specialised teams with single agents based on the functional correctness and requirement coverage of the delivered software \cite{dong2024selfcollab,zeng2025e2edevbench}; a controlled comparison across 260 configurations evaluates each one using benchmark-specific measures of task performance \cite{kim2026outgrow}. Cost studies measure the tokens agentic workflows consume and how cost varies with task difficulty, and argue that cost belongs beside accuracy in agent evaluation \cite{kapoor2025agents,wang2025efficient,bai2026tokens}. Studies of failure catalogue the ways multi-agent pipelines break, from mis-specification to inter-agent misunderstanding: MAST derives fourteen failure modes from execution traces \cite{cemri2025mast}, attribution work asks which agent and which step caused a failed run \cite{zhang2025whowhen}, and a study of code-generation teams attributes three quarters of the failures it analyses to the boundary between planner and coder \cite{lyu2025plannercoder}. All of this looks at finished run: a grade, a bill, or a post-mortem of the transcript.

A second line of work designs the communication structure in advance: which agents may talk to which, in what order, and through what roles. ChatDev and MetaGPT assign fixed roles and pass work along a prescribed order \cite{qian2024chatdev,hong2024metagpt}; multi-agent debate fixes who reads whom and for how many rounds \cite{du2024debate}; MacNet arranges agents into chains, stars, trees, and meshes and compares the arrangements \cite{qian2025scaling}; GPTSwarm and AgentPrune treat the communication graph itself as a quantity to optimise or prune \cite{zhuge2024gptswarm,zhang2024agentprune}. In every case the topology is an input: the system fixes or searches the structure and then scores the outcome, so the coordination that happens inside the structure goes unmeasured. When communication is examined at all, it appears as raw message logs or token counts, and the structure, the timing, and the file activity that carry much of the coordination stay out of view. Our experiment leaves the channel open: any agent may message any other under every configuration, and the topology is a result the instrument recovers (Section~\ref{sec:topology}).

What is missing is an instrument: a way to observe the coordination structure that emerges during a run, on the same footing for messages and for shared artefacts, and comparable across team sizes, configurations, and frameworks. Network analysis provides the natural language for this, and it has a long record in software engineering. Developer networks mined from version histories predict failure-prone components \cite{pinzger2008developer,meneely2008predicting}, and combining the contribution network with the technical dependency network predicts them better than either alone \cite{bird2009puttingit}. The socio-technical congruence tradition compares the coordination a project's dependencies demand with the coordination its records show \cite{cataldo2006identification,cataldo2008congruence}. Longitudinal network studies expose team structure itself: developer coordination networks in large open-source projects organise into stable cores and loosely connected peripheries as they evolve \cite{joblin2017trends,joblin2017core}. All of these networks are recovered from repository artefacts after the fact, at the granularity the commit record allows; even time-stamped co-editing networks are reconstructed from Git histories once the commits exist \cite{gote2019git2net}. Multi-agent runs permit something stronger: every message, write, and read made through the runtime's tools is a logged event, so the temporal network of a collaboration can be recorded directly from the trace as it happens.

This paper supplies that instrument. It differs from output-based evaluation in what it measures (the coordination itself), from failure analysis in timing (the run is recorded as it unfolds, before its outcome is known), from topology design in direction (the network is an observed outcome), and from mined developer networks in resolution (every edge is an exact, timestamped event). The closest prior threads, on span of control and group-size limits in human organisations \cite{graicunas1933,miller1956,dunbar1992,zhou2005}, return in Section~\ref{sec:conclusion}, where the measured limits on sustained coordination invite the comparison.
\section{Conclusion}
\label{sec:conclusion}

Multi-agent AI coding systems are evaluated by their outputs, but they succeed or fail by their coordination, and the coordination is invisible in the output. This paper's contribution is to make it a measured object: a temporal network in which agents and files are both nodes and every message, write, and read logged by the runtime's tools is a timestamped, costed edge. The graph is a direct record of that tool-mediated activity, cheap to collect from existing logs, and comparable across team sizes, configurations, and frameworks.

Measured this way, the three configuration choices a developer faces come out differently from what output-level evaluation suggests. The quadratic cost of communication is real but shallow: it is an opening handshake, made once, after which each agent sustains a few channels, no hub forms, and the largest teams switch increasingly to broadcast and messaging growth levels off between eight and sixteen agents. Files, usually treated as outputs, are the team's one-to-many channel, and requiring their use cuts output tokens by about 42\% at eight agents on message-heavy work, while adding cost on chain-shaped work whose files already carry the coordination. And leadership assigned by prompt is nominal only: it creates no structure and buys no reliable success, and a sealed replication of the eight-agent conflict cells leaves flat and coordinator teams level under every file policy. The common thread is that what the prompt declares and what the interaction structure builds can differ, and the measured outcomes follow the second.

Beneath those choices, the shape of the coordination network is set by the task. Left to organise themselves, the teams do not converge on a single coordination span. The distributed task, where everyone reconciles one shared specification, builds a dense, tightly clustered graph that rides the all-to-all line; the chained task, where each agent agrees interfaces only with its neighbours, stays sparse, down to a mean degree of 0.28 against a clique of fifteen at sixteen agents, where scarcely any named-message network forms at all. Neither has a hub. In our experiments, the structure a team builds is a property of the work it is given, read directly off the graph.

The two collections also reveal the reliability of the measurements. Under one pinned model, the chained task's measurements replicate across sessions to two decimal places while the distributed task's exponent moves from 1.76 to 2.44, so a configuration is a distribution of runs, and single-run evaluations of multi-agent systems stand on a sample of size one. And the measurements see what grading cannot: an eight-step calculation split one step per agent failed every run on a rounding convention that sat between two agents and belonged to neither, discussed every time and never settled. Decomposition creates interfaces between agents, each interface needs an owner, and the graph shows which have none.

The instrument also caught a behaviour that grading hides. The main runs were not confined to their workspace, and the teams used the access: they opened the hidden grading suite in 234 runs and the reference solution in 77. A sealed replication of the load-bearing cells, with decoys where the real files had sat, both confirmed the two findings it re-tested under containment, the coordinator null and the message-to-file substitution, and measured the behaviour directly: in four fifths of the sealed runs the agents reached for the grading suite even though it returned nothing, an unprompted search for the answer key that only a record of coordination, and not a grade, can see.

There is a resonance here with a long literature on the limits of human coordination. Spans of workable oversight have been placed at around five direct reports by Graicunas \cite{graicunas1933} and near seven items by Miller \cite{miller1956}, while Dunbar links group size to limits on the relationships that can be monitored \cite{dunbar1992}, and human group sizes cluster near five and fifteen in Zhou et al.'s analysis \cite{zhou2005}. Our agent teams sustain a limited number of one-to-one pushes, roughly two to five peers per agent by the directed out-degree of Section~\ref{sec:topology}, in the range of those figures, and in the sixteen-step arm messaging growth halts between eight and sixteen agents, which invites the comparison. The topology result cautions against reading it as a shared cognitive span. Left free, the agents do not settle on one span at all: where the task calls for it they build a dense near-clique, and where it does not they stay sparse. Human hierarchies may owe as much to the cost of adding people and the distribution of authority as to any limit on coordination itself; here, adding agents still carries coordination cost, but formal authority is minimal, and the pyramid does not return. What remains and still binds is the cost of coordination. We claim no shared mechanism, and our design cannot place the growth break more precisely than the eight-to-sixteen interval, but if coordination structure is subject to similar economics wherever it arises, machine teams are a way to study those economics with an exactness that is difficult to achieve in human organisations.

The framework, the pipeline, and all 1{,}902 main-collection runs, together with the 244 sealed replication runs, are released for replication. The next steps are the ones the instrument makes possible: naturalistic tasks from real repositories, other models and runtimes, the handshake as an early-warning signal for failing runs, and the cross-team setting where separate agent teams share no message bus but do share a repository, so the files, already first-class nodes here, carry all the coordination there is to see.

\section*{Data Availability}
The replication package, containing all 1{,}902 runs as CSV datasets, the task generators, the instrumentation pipeline, and every analysis script behind the numbers and figures in this paper, is released at \url{https://github.com/giuseppedestefanis/when-agents-coordinate}. The 244 sealed replication runs of Section~\ref{sec:containment} are released alongside it, together with the seal implementation and the script that classifies each read by location. The pre-registration records are included, so each committed prediction and decision rule of Appendix~\ref{app:stats} can be checked against the analysis it governs, including the top-up decision plan and the qualifications on H2 and H3; the decision-rule bodies are the committed versions, and the package notes where a scaling-arm record's status header was updated after collection. Each reported statistic can be regenerated from the released data without rerunning any agent.

\bibliographystyle{plain}
\bibliography{references}

\appendix
\section{Pre-Committed Hypotheses and Statistical Detail}
\label{app:stats}

\subsection{The eight hypotheses}

The pre-registrations released with the package are the Experiment 2 plan, which states H1 and H3--H6 with their tests, and the two scaling-arm commitments, which state H7 and H8. The Experiment 1 plan released alongside them pre-registers only the top-up decision above, so H2 is not a fully pre-registered hypothesis: its predicted direction was fixed from the Experiment 1 pilot, but its Fisher test and its correction set were specified after the schedule had run, before the inferential analysis. H3 carries a separate qualification: the Experiment 2 plan pre-registers a within-Experiment-2 Fisher test of the flat condition (labelled \texttt{peer} in the pre-registration) against the coordinator condition (labelled \texttt{orchestrator}) on the conflicting split, which is inconclusive, whereas the cross-experiment interaction reported in Section~\ref{sec:leadership} ($p = 0.24$) is an exploratory test the plan does not specify. The hypotheses committed in full, in both prediction and pre-registered test, are therefore H1 and H4--H8; H2 and H3 carry the qualifications above, and both are reported as inconclusive or null. One note for the reader who opens the released plan: the Experiment 2 plan uses its own internal numbering, in which H2 and H7 are different hypotheses (a topology-distribution prediction and a corroborative footprint check) that do not enter the paper's set; the package includes a crosswalk from that numbering to the numbering used here. Where the body reports a hypothesis as confirmed or not supported, it refers to the statement and test recorded here.

\begin{table}[h]
  \centering
  \caption{The eight hypotheses and their outcomes. H1 and H4--H8 are pre-registered in both prediction and test; H2 and H3 carry qualifications (see text).}
  \label{tab:hypotheses}
  \footnotesize
  \begin{tabular}{@{}l p{4.9cm} p{2.5cm} p{4.6cm}@{}}
    \toprule
    & hypothesis, in one phrase & committed before & outcome (where) \\
    \midrule
    H1 & chained-task messaging scales as $n^2$ & Experiment 2 collection & confirmed (\S\ref{sec:handshake}) \\
    H2 & a coordinator helps on conflicting tasks & direction before collection, test after & directional at $n{=}4$; the $n{=}8$ reversal does not survive the seal and is treated as null (\S\ref{sec:leadership}, \S\ref{sec:containment}) \\
    H3 & the coordinator effect differs across experiments & Experiment 2 collection (within-experiment Fisher; cross-experiment test exploratory) & inconclusive (\S\ref{sec:leadership}) \\
    H4 & mandatory policy adds file coordination & Experiment 2 collection & confirmed (\S\ref{sec:files}) \\
    H5 & chained-task addressing is less peer-directed & Experiment 2 collection & confirmed (below) \\
    H6 & a shared constant reduces peer-directed addressing & Experiment 2 collection & not supported (below) \\
    H7 & the $n{=}4$ break is a coordination property & 8-step arm & directional, power-limited (\S\ref{sec:handshake}) \\
    H8 & messaging growth halts between 8 and 16 agents & 16-step arm & confirmed (\S\ref{sec:handshake}) \\
    \bottomrule
  \end{tabular}
\end{table}

Two hypotheses are not discussed in the body. H5: addressing differs sharply between the experiments. At the matched four-agent cell (collection B, allowed policy, clean split), the per-run mean share of messages that name a specific peer is 97.1\% in Experiment 1 against 60.7\% in Experiment 2, a 36-point gap in the pre-committed direction. H6: a constant shared by every chain step was predicted to push agents towards broadcasting it, reducing the share of messages aimed at one named peer by at least fifteen points; the observed drop is 4.1 points (28.3\% to 24.2\%, $p = 0.23$), in the predicted direction but far short of the margin. The teams resolved the shared dependency by writing the constant into a file, which is Finding 2 appearing where a messaging effect was predicted.

\subsection{Tests and corrections}

The unit of analysis is one run: one graph, one binary outcome. Binary success rates carry Clopper--Pearson exact 95\% intervals \cite{clopper1934}. Success contrasts use Fisher's exact test \cite{fisher1934smrw}; continuous contrasts use the Mann--Whitney rank test \cite{mann1947}; neither assumes normality. Where a question tests a set of related contrasts, the set is corrected with the Benjamini--Hochberg procedure \cite{benjamini1995} and the corrected values are reported as $p_{\mathrm{BH}}$. Scaling exponents are least-squares slopes of $\log(\text{count})$ on $\log(n)$ over per-run values, with 95\% confidence intervals. The one exception is the pre-registered H1 test, which by its plan regresses over the three cell means; it gives the same slope with a wider interval, and both forms are reported in Section~\ref{sec:handshake}. The scaling arms use a piecewise version with the knot fixed in advance ($n{=}4$ for the eight-step arm, $n{=}8$ for the sixteen-step arm) and report the two segment slopes and their difference $\Delta$.

\subsection{The sustained undirected graph and its topology}
\label{app:topology}

Section~\ref{sec:topology} reads the shape of a run's coordination from its sustained undirected graph. For one run, count the direct messages sent in each ordered direction between every pair of agents. Two agents are joined by an undirected edge when at least one of the two directions carried two or more messages, so single greetings are dropped and a link marks a channel the pair used more than once. Only the configured agents \texttt{agent-1} to \texttt{agent-$N$} are admitted as endpoints, so a hallucinated recipient never enters the graph. The \emph{mean degree} of a run is the sum of the agents' degrees divided by the full team size $N$, so an agent that sustains no channel counts as degree zero and stays in the denominator; a run that builds no sustained named-message network therefore has mean degree zero. We report the mean of this quantity over \emph{every} run in the cell, silent runs included, against the clique line $n-1$, the degree every agent would have if the team were fully connected. Averaging over every configured agent and every run, including the silent ones, is what keeps the degree an honest per-agent expectation; the earlier of these two choices matters most on the sparse sixteen-agent cell, where most runs build no network at all. The \emph{global clustering coefficient} (transitivity) of a run is the number of closed triples divided by the number of connected triples in the same graph: for every agent, over each unordered pair of its partners, the pair is a connected triple, and a closed one when those two partners are themselves joined. It is 1 when the live partners form a single interconnected group and 0 when no two partners of any agent are joined. Both quantities are computed by \texttt{precompute\_topology\_scaling.py} in the released package, over the flat cells (both collection sessions pooled) under the allowed policy and clean split, at two, four, and eight agents for each task, with the sixteen-step chain adding points at eight and sixteen agents, the latter the only sixteen-agent data. The directed out-degree of Figure~\ref{fig:sustained} (the \texttt{eff\_team\_t} measure in the released code) is a different measurement on the same messages: for each agent the number of distinct peers it \emph{sends} at least $t$ messages to, averaged across all agents; the count is directed and differs from the degree of the undirected graph.

\subsection{Design checks}

Repeated runs of the same configuration vary widely \cite{bjarnason2026randomness,mustahsan2025stochasticity}, in agentic coding by up to thirty-fold in token consumption \cite{bai2026tokens}, and every main-matrix cell carries ten repetitions. A pre-registered rule flags a cell for topping up to twenty when its precision is low, and applied to the released data it flags 93 cells across the two experiments (48 of the 85 in Experiment 1, 45 of the 87 in Experiment 2), 25 of them on the outcome-precision condition and the rest on graph-statistic variance. The flags divide roughly evenly by team size, 35 at two agents, 36 at four, and 22 at eight. We topped up none of them, which is a deviation from the rule as written, made deliberately and recorded at the time. Topping the flagged cells up to twenty runs would have sharpened their estimates, the graph-statistic means as much as the success rates, by the usual factor of about $\sqrt{2}$; we judged it unnecessary because the confirmatory success contrasts pool across cells, so no single cell decides them, and because many of the graph-statistic flags were triggered by a metric other than the one a headline result uses for that cell, a cell flagged on file-read variance, say, whose headline reading is message count or sustained degree. The cost of that judgement is that the affected cell-level graph estimates remain less precise than the rule intended. A few hypotheses do read specific flagged cells, among them H1, H5, and H6, and their intervals should be read with that reduced precision in mind. The strict alternative of topping up every flagged cell was considered and set aside, and we record the departure here. The one deviation in the other direction is the sixteen-step scaling arm, pre-registered at twenty runs per allowed cell to tighten the plateau test of hypothesis H8. Two ablations defend the design. First, the forbidden policy is a prompt clause; re-running the worst-affected cell with the workspace locked at the filesystem level produced the same workspaces and the same outcomes, so the prompt form is sound. Second, the pilot's dominant failure traced to an ambiguous return-type clause in one specification; the clause was tightened and the worst cell re-validated (three of three passes) before the full schedules ran, leaving a cleaner outcome variable. A within-batch regression of messaging on run order shows no drift, so batches are internally stable; the between-session drift is measured and reported in Section~\ref{sec:reliability}.

\end{document}